\documentclass[a4paper,fleqn]{cas-dc}
\let\printorcid\relax % Remove ORCID footnote
\usepackage{float}
\usepackage[sort, authoryear,longnamesfirst]{natbib}
{}
{}
{}
\usepackage{graphicx}
\usepackage{algorithm}
\usepackage{algpseudocode}
\usepackage{caption}
\usepackage{subcaption}
\usepackage{amsmath}
\usepackage{amssymb}
\usepackage{eqparbox}
\newdimen{\algindent}
\algnewcommand\LeftComment[2]{%
	\hspace{#1\algindent}$\triangleright$ \eqparbox{COMMENT}{#2} \hfill %
}
\usepackage{mathtools, nccmath}

\usepackage{tikz}
\usetikzlibrary{shapes.geometric, arrows}

\tikzstyle{startstop} = [rectangle, rounded corners, 
minimum width=3cm, 
minimum height=1cm,
text centered, 
draw=black, 
fill=red!30]

\tikzstyle{io} = [trapezium, 
trapezium stretches=true, % A later addition
trapezium left angle=70, 
trapezium right angle=110, 
minimum width=3cm, 
minimum height=1cm, text centered, 
draw=black, fill=blue!30]

\tikzstyle{process} = [rectangle, 
minimum width=3cm, 
minimum height=1cm, 
text centered, 
text width=3cm, 
draw=black, 
fill=orange!30]

\tikzstyle{decision} = [diamond, 
minimum width=3cm, 
minimum height=1cm, 
text centered, 
draw=black, 
fill=green!30]
\tikzstyle{arrow} = [thick,->,>=stealth]

\def\tsc#1{\csdef{#1}{\textsc{\lowercase{#1}}\xspace}}
\tsc{WGM}
\tsc{QE}
\usepackage{calc}
\usepackage{ifthen}
\usepackage{hyperref}

\newcounter{captionedequationset} %numbering
\newdimen\captionlength
\newcommand{\captionedequationset}[1]{
	\refstepcounter{captionedequationset}% Step counter
	\setlength{\captionlength}{\widthof{#1}} %
	\addtolength{\captionlength}{\widthof{Equation set~\thecaptionedequationset: }}
	\ifthenelse{\lengthtest{\captionlength < \linewidth }} %
	{\begin{center}
			Equation set~\thecaptionedequationset: #1
	\end{center}} 
	{ \begin{flushleft} 
			Equation set~\thecaptionedequationset: #1 %
\end{flushleft}}}

\usepackage{stackengine}

\usepackage[inline,shortlabels]{enumitem}
\usepackage{tabto}
\usepackage{booktabs}
\begin{document}
\let\WriteBookmarks\relax
\def\floatpagepagefraction{1}
\def\textpagefraction{.001}

\shorttitle{SimGen}    
%\shortauthors{Achara et al.} 

\title [mode = title]{Revealing the Underlying Patterns: Investigating Dataset Similarity, Performance, and Generalization}  

%\tnotemark[<tnote number>] 

%\tnotetext[<tnote number>]{<tnote text>} 

%       orcid=---,

\author[1]{Akshit Achara}

%\cormark[1]


\ead{f2016953p@alumni.bits-pilani.ac.in}

\affiliation[1]{organization={GE Research},
	city={Bangalore},
	country={India}}

\author[1]{Ram Krishna Pandey}

% Footnote of the second author

% Email id of the second author
\ead{ramp@alum.iisc.ac.in}



\begin{abstract}
		Supervised deep learning models require significant amount of labeled data to achieve an acceptable performance on a specific task. However, when tested on unseen data, the models may not perform well. Therefore, the models need to be trained with additional and varying labeled data to improve the generalization. In this work, our goal is to understand the models, their performance and generalization. We establish image-image, dataset-dataset, and image-dataset distances to gain insights into the model's behavior. Our proposed distance metric when combined with model performance can help in selecting an appropriate model/architecture from a pool of candidate architectures. We have shown that the generalization of these models can be improved by only adding a small number of unseen images (say 1, 3 or 7) into the training set. Our proposed approach reduces training and annotation costs while providing an estimate of model performance on unseen data in dynamic environments.
	
\end{abstract}


%	\item We proposed a distance metric that can be used for inter and intra dataset comparison.
%	\item The distances can be related to performance of the models.
%	\item Models can be directed towards generalization with very few images.
%	\item Model behaviour can be explained using the proposed methods.

% Keywords

\maketitle
\newpage

% Main text
\section{Introduction and Related Work}
\label{intro}
Deep Learning tasks like segmentation are commonly performed by using supervised techniques. This requires labeled data for consumption by the model. If the model is not trained on diverse data, it can overfit and may not perform well on unseen data. It might also be the case that the model is not expressive enough to handle the variations in the training data thereby showing poor performance. So, the model needs to be trained with additional data points from unseen data to achieve better performance. This boils down to the question: "how much additional labeled data is required?".

Therefore, an analysis of model performance and behaviour on test and unseen data is required. Generally, test datasets have similar distribution to the training datasets but unseen datasets may or may not have similar distribution. Hence, it requires us to obtain a metric that can tell how far these unseen datasets are from the training dataset. 
In order to accomplish this task, we need to compare datasets by systematically evaluating the images in one dataset against all images in the other dataset. Comparison of the raw images is computationally expensive and is not robust to the variations such as lightning, illumination, orientation, etc. Therefore, we extract feature vectors as the representative of images.

Classical approaches utilizing structural similarity \ref{ssim} and histogram of oriented gradients \ref{hog} have been used to find descriptors and keypoints from the images. These methods have been commonly used to distinguish noisy images from a set of images.

Deep learning techniques based on Siamese networks \ref{siamese} have been commonly used for computing image similarity. In \ref{siamesematching}, the authors utilize a CNN-based Siamese architecture with contrastive loss to compute Euclidean distance between the images using feature vectors. However, this approach is constrained by the requirement for labeled matching and non-matching pairs of images.

While the aforementioned methods are focused on image similarity, in \ref{otdd}, the authors introduced an optimal transport solution to compute the distance between the datasets. However, this approach considers both labels and images within a dataset, which doesn't align with our objective of accommodating analysis for unseen datasets.

The subsequent techniques contribute to the advancement of generalization and data selection. In \ref{dmlgen}, the authors conduct a theoretical analysis of generalization error bounds of deep metric learning (DML) and introduce ADroDML, an adaptive dropout technique validated through experiments, but it is limited by the need for labelled data pairs or triplets during training.
In \ref{dmlnet}, the authors propose an open world image segmentation framework to detect in-distribution and out-of-distribution (OOD) objects, along with a few-shot learning module for OOD object adaptation. However, this work primarily emphasizes on class-wise adaptation, whereas our approach does not depend on class labels and remains agnostic to the number of classes in unseen data.
Techniques proposed in \ref{visgenunets} provide insights into the generalization capabilities using of UNets using metrics like roughness without requiring ground truth annotations. However, the authors study layerwise contributions of a UNet for segmentation and a few CNN models for classification.
\ref{simperf} introduces a novel concept of similarity between training and unseen data, investigating its correlation with the F-score of an FCN classifier. The proposed landscape metrics for similarity are however focused on urban studies. Meanwhile, \ref{eqperf} demonstrates that merely increasing the number of training examples may not necessarily enhance model performance, emphasizing the significance of a well-designed data selection strategy. Additionally, in \ref{ugenvis}, the authors provide valuable insights into neural network generalization, offering visualization from the perspectives of optimization and loss.

In the broader context of research on model generalization, data similarity, and selection, our work stands out by introducing novel distance metrics. We first establish the foundational strength of these metrics by initially utilizing images from entirely different scenes, gradually transitioning to an in-depth analysis of similar scenes. This analysis is integral to our investigation as we correlate these metrics with the performance and generalization of models featuring different architectures.

Furthermore, we conduct experiments to understand how these selected models behave across various domains, thereby assessing their applicability in diverse contexts. Our experimentation also showcases the adaptability of the model with minimal data on different datasets. To ensure the rigor of our study, we have selected publicly available datasets for our experiments and analysis.

\subsection{Contributions}
\label{contributions}
Our main contributions are as follows:
\begin{enumerate}[a)]
\item We proposed a distance metric to obtain the distances between datasets ($O^{dist}$) and between images and datasets ($I^{dist}$) that can be related to the performance of the model.
\item If the F-score is consistent (F-score vs $I^{dist}$ curve is a line parallel to the x-axis) across all the unseen images (see figure~\ref{fig:dc_if}), model need not be finetuned saving energy and labeling cost.
\item We have shown that selecting a few images from an unseen dataset, can significantly boost its performance thereby improving the generalization and reducing annotation cost (see figures~\ref{fig:unet_if}, ~\ref{fig:adsam_if}, ~\ref{fig:udcadforest} and~\ref{fig:cfd}).
\item Our study can give a relative comparison of models and their tradeoffs that can possibly help in selecting the most suitable model based on the requirements (see section~\ref{observations} for more details).
\item We have found that models like segment anything \ref{sam} and adapted segment anything \ref{adsam} require additional data to perform well across multiple unseen datasets for a specific task (see section~\ref{samcrack} in appendix and figure~\ref{fig:cfd} for details).
\end{enumerate}

\section{Datasets}
\label{datasets}
\subsection{Crack Datasets}
\label{crackdatasets}
\begin{enumerate}[a)]
	\item \textbf{CrackTree260} ($P$):
	It contains 260 road pavement images. 
	\item \textbf{CrackLS315} ($S_1$):
	It contains 315 road pavement images of size $512\times512$. 
	\item \textbf{CRKWH100} ($S_2$):
	It contains 100 road pavement images  of size $512\times512$. 
	\item \textbf{GAPS} ($S_3$):
	It contains 509 images of size $448\times448$ selected from the kaggle crack segmentation dataset.
	\item \textbf{FOREST} ($S_4$):
	It contains 118 images of size $448\times448$ selected from kaggle crack segmentation dataset.
\end{enumerate}
\subsection{Non-Crack Datasets}
\begin{enumerate}[a)]
	\item \textbf{PASCAL-VOC} ($S_5$):
	It is a dataset \ref{pascalvoc} with 20 classes and multiple tasks like Classification, Detection, Segmentation and Action Classification (10 classes) are defined on it. We randomly selected 500 images from the dataset to create a fixed new dataset ($S_5$) for all the experiments in this study.
	\item \textbf{BSDS500} ($S_6$):
	It is a dataset \ref{bsds500} that contains 500 images of size $421\times321$ commonly used for benchmarking on segmentation and boundary detection tasks.
\end{enumerate}
$P$, $S_1$ and $S_2$ were obtained from \ref{deepcrack} and, $S_3$ \ref{gaps} and $S_4$ were obtained from \ref{kagglecrack}. In this work, the dataset $P$ is considered as the primary dataset and the datasets $S_i, i \in \{1,2,3,4,5,6\}$ will be referred as the secondary datasets ($S$ will be interchangeably used with $S_i$).  We resized all the images to $448\times448$ for all the experiments.

\section{Methodology}
\label{methodology}
\subsection{Image Representation}
\label{imagerepresentation}
We considered multiple models to get feature vectors from the images, namely Segment Anything Model \ref{sam} ("base" version is used in this study), CLIPSeg \ref{clipseg} (segmentation model to perform image segmentation using text and image prompts), EfficientNet \ref{efficientnet} model pretrained on imagenet \ref{imagenet} dataset, DeepCrack \ref{deepcrack} which is a specialized model for crack segmentation (hereafter referred as DC), UNet${++}$ \ref{unet++} initialized with an EfficientNet backbone (hereafter referred as UNet$^{++}$) and an adapted version of SAM that can be trained on custom datasets \ref{adsam} (hereafter referred as ADSAM).

In \ref{clip}, the authors propose a contrastive language-image pre-training (CLIP) model to find the most relevant text given an image. In this work, we use the modified CLIP (CLIPSeg \ref{clipseg}) model where a decoder is added to perform segmentation tasks.
We obtain the feature vectors by concatenating the outputs of all hidden layers of the decoder by giving input text prompts along with the images to the CLIPSeg model. The checkpoint used can be seen here\footnote{\url{https://huggingface.co/CIDAS/clipseg-rd64-refined}}.

We extract the feature vectors from each model to obtain a meaningful high dimensional feature representation of images.

\begin{table}
	\begin{tabular}{lp{5cm}}
		\toprule
		Model & Feature Vector \\
		\midrule
		SAM	  & The image embeddings from the image encoder ( MAE pre-trained Vision Transformer (ViT)\ref{mae}) of the SAM\ref{sam}.\\
		
		CLIPSeg & The concatenated hidden states' outputs from the CLIP decoder~\ref{clipseg} with the input text prompt \textbf{"line structures"}.\\
		
		ENet & The last layer(before the classification head) of a pretrained EfficientNet~\ref{efficientnet} Model on imagenet. \\
		
		DC & The concatenation of all the downsampling
			layers' outputs of the encoder as shown in the architecture of DC~\ref{deepcrack}.\\

		$UNet^{++}$ & The concatenation of all the skip connections from the EfficientNet  encoder \ref{efficientnet} used as backbone in $UNet^{++}$.\\
		
		$ADSAM$ & The image embeddings from the image encoder discussed in \ref{adsam}. \\
		\bottomrule
	\end{tabular}
	\caption{The table shows the model and the details of the corresponding feature vector extracted from it. The details of the models can be seen in section~\ref{imagerepresentation}.}
	\label{featurevector}
\end{table}

\subsection{Distance Computation}
\label{distancecomp}
The high dimensional feature vectors obtained from the models are  projected into a low dimensional representation using PCA to capture the distinguishing features while reducing the complexity of comparing multiple images and datasets.

We considered feature vectors extracted from all images of two datasets at a time for projection (into a low dimensional space) where one dataset set is always $P$ and the other is one of the secondary datasets $S$. We use 25 principal components together for the study as we observe stability in distance computation (see section~\ref{dimselection} in appendix for more details).

We compute the pairwise distances between the images of P and $S$ using the low dimensional vectors to obtain a distance matrix as shown in equation~\ref{pairwisedist}. Sum of each row of the pairwise distance matrix represents image-dataset distance($I^{dist}$) i.e. the distance of each image in $S$ from all of $P$, and mean of all the rows taken together represent dataset-dataset distance($O^{dist}$) i.e. the overall distance of $S$ from $P$. See equation set~\ref{imgdist} for more details.
In essence, taking any two datasets in consideration namely, primary and secondary, O$^{dist}(S, P)$ represents the distance between the two datasets and I$^{dist}(S, P)$ represents the distance between an image of secondary dataset from the entire primary dataset. I$^{dist}$ is image specific and varies based on the selected image.
$O^{dist}$ can be computed for datasets of different sizes since it is equal to the mean $I^{dist}$ i.e. the mean of distances of each image of the secondary dataset from the entire primary dataset.

\begin{figure*}[htpb!]
	\centering
	\includegraphics[width=0.90\textwidth,height=0.22\textheight]{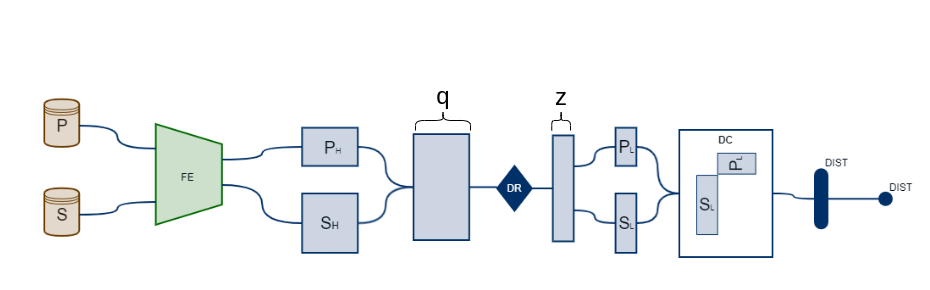}
	\caption{The figure shows the distance computation process discussed in the section~\ref{distancecomp}. FE is the feature extractor/model from which high dimensional feature vectors $P_H$ and $S_H$ are obtained. The DR (dimensionality reduction) using PCA results in low dimensional representations $P_L$ and $S_L$ from q to z dimensions of the (n+m) images. Finally, the $I^{dist}$ and $O^{dist}$ are computed using the paiwise distance matrix shown in the DC(distance computation) block.}
	\label{fig:distcompprocess}
\end{figure*}

Since the proposed distance metrics are only utilizing images and not labels, the distance computation can be used for multiple other tasks like classification, object detection and even for the other data types. However, the feature vectors should be extracted based on the input and model architecture such that the input features are captured. The entire computation process can be seen in the figure~\ref{fig:distcompprocess}.

\begin{align}
	\centering
\begin{split}
		F(x) = y; x \in R^{h \times w}, y \in R^q
\end{split}
\end{align}
\begingroup\vspace*{-\baselineskip}
\captionedequationset{The feature extractor ($F$) maps the input image to a feature vector. Here, $h \times w$ represents the size of the input image i.e. $448\times 448$.}
\vspace*{\baselineskip}\endgroup

\begin{align}
	\centering
	\begin{split}
		P_H = \begin{Bmatrix}
			F_1(x_P)\\
			F_2(x_P)\\
			\vdots \\
			\vdots \\
			F_n(x_P) \\
		\end{Bmatrix}_{n\times q}; x_P = \{x \vert x \in P\}\\
		S_H = \begin{Bmatrix}
			F_1(x_S)\\
			F_2(x_S)\\
			\vdots \\
			\vdots \\
			F_m(x_S) \\
		\end{Bmatrix}_{m\times q};x_S = \{x \vert x \in S\}
	\end{split}
\end{align}
\begingroup\vspace*{-\baselineskip}
\captionedequationset{$P_H$ and $S_H$ represent the feature vectors of primary and secondary datasets respectively where each q-dimensional row represents a feature vector of an image. Here, $n$ is the number of images in the primary dataset and $m$ is the number of images in the secondary dataset.}
\vspace*{\baselineskip}\endgroup

\begin{align}
	\centering
	\begin{split}
		X = \begin{Bmatrix}
			P_H^{\top} \lvert S_H^{\top}
		\end{Bmatrix}^\top_{(n+m) \times q}\\
		X = \begin{Bmatrix}
			X_1 \\
			X_2 \\
			\vdots \\
			\vdots \\
			X_{(n+m)}
		\end{Bmatrix}_{(n+m) \times q} \\
		\bar{X} = \frac{1}{(n+m)} \sum_{j=0}^{(n+m)} X_j \\
		A = X - \bar{X} \\
	\end{split}
\end{align}
\begingroup\vspace*{-\baselineskip}
\captionedequationset{$A$ in the centered input for the PCA which is created by concatenating the transposed $P_H$ and $S_H$ matrices and transposing them again to get a matrix of $q$ columns that will be converted to a low dimensional representation.}
\vspace*{\baselineskip}\endgroup

\begin{align}
	\centering
	\begin{split}
		A = \underset{(n+m)\times (m+n)}{\mathrm{U}}\underset{(n+m)\times q}{\mathrm{\sum}}\underset{q\times q}{\mathrm{V^\top}} \\
		A^\top A = V\sum\nolimits^\top U^\top U \sum V^\top\\
			= V \sum\nolimits^\top \sum V^\top \\
			= V D V^\top \\ 
		%A^\top A V = V D
	\end{split}
\end{align}
\begingroup\vspace*{-\baselineskip}
\captionedequationset{Here, $U$ and $V$ are orthogonal matrices i.e. $U^\top U = V^\top V = I$. $V$ represents the eigen vectors and $D$ represents the eigenvalues.}
\vspace*{\baselineskip}\endgroup

\begin{align}
	\begin{split}
	V_{(n+m)} = \begin{Bmatrix}
		v_{11}& v_{12}& ...& v_{1q} \\
		v_{21}& v_{22}& ...& v_{2q} \\
		v_{31}& v_{32}& ...& v_{3q} \\
		\vdots & \vdots & ... & \vdots \\
		\vdots & \vdots & ... & \vdots \\
		v_{1q}& v_{2q}& ...& v_{qq} \\
	\end{Bmatrix} \\
	V_{z} = \begin{Bmatrix}
		v_{11}& v_{12}& ...& v_{1z} \\
		v_{21}& v_{22}& ...& v_{2z} \\
		v_{31}& v_{32}& ...& v_{3z} \\
		\vdots & \vdots & ... & \vdots \\
		\vdots & \vdots & ... & \vdots \\
		v_{q1}& v_{q2}& ...& v_{qz} \\
	\end{Bmatrix} \\
	\underset{(n+m)\times z}{\mathrm{Y}} = A V_z \\
	Y = \begin{Bmatrix}
		P_L^\top \lvert S_L^\top
	\end{Bmatrix}^\top_{(n+m) \times z}
	\end{split}
\end{align}

\begingroup\vspace*{-\baselineskip}
\captionedequationset{$Y$ consists of a low dimensional representation of  the feature vectors of $P$ and $S$ obtained using PCA (implementation taken from here \ref{scikit}). $z$ is the number of principal components in the low dimensional representation of the images.}
\vspace*{\baselineskip}\endgroup

\begin{align}
	\centering
	\begin{split}
	D(S, P) = 
	\begin{Bmatrix}
		d(s_{1}, p_1) & d(s_{1}, p_2) & ... &d(s_{1}, p_n)\\
		d(s_{2}, \rho_1) & d(s_{2}, p_2) & ... &d(s_{2}, p_n)\\
		d(s_{3}, p_1) & d(s_{3}, p_2) & ... &d(s_{3}, p_n)\\
		\vdots & \vdots & ... & \vdots \\
		\vdots & \vdots & ... & \vdots \\
		d(s_{m}, p_1) & d(s_{m}, p_2) & ... &d(s_{m}, p_n)
	\end{Bmatrix}\\
	d(a, b) = \lVert a - b \rVert_2	
\end{split} 
\label{pairwisedist}
\end{align}

\begingroup\vspace*{-\baselineskip}
\captionedequationset{$D(S, P)$ refers to the pairwise distance matrix where each value in the matrix is the Euclidean distance between each image of $S$ and $P$. $s_j$ represents the low dimensional feature vector from $S_L$ and $p_k$ represents the low dimensional feature vector from $P_L$. Here, $j \in \{1,2,..,m\}$ and $k \in \{1,2,...,n\}$. From here onwards, we can consider $S_L$ and $P_L$ are represented by $S$ and $P$ in the context of distance computation.}
\vspace*{\baselineskip}\endgroup

\begin{align}
	\centering
	\begin{split}
		I^{dist}(S, P) = \begin{Bmatrix}
			\sum_{j=1}^{n} d(s_{1}, p_j) \\
			\sum_{j=1}^{n} d(s_{2}, p_j) \\
			\sum_{j=1}^{n} d(s_{3}, p_j) \\
			\vdots \\
			\vdots \\
			\sum_{j=1}^{n} d(s_{m}, p_j) \\	
		\end{Bmatrix}\\
		I^{dist}(S, P) = \begin{Bmatrix}
			I_1^{dist}(S, P) \\
			I_2^{dist}(S, P) \\
			\vdots \\
			\vdots \\
			I_m^{dist}(S, P) \\
		\end{Bmatrix} \\
	O^{dist}(S, P) = \frac{1}{m} \times \sum_{j=1}^{m} I_j(S, P) \\
	\end{split}
	\label{imgdist}
\end{align}

\begingroup\vspace*{-\baselineskip}
\captionedequationset{$I^{dist}$ represents the distance of each image in $S$ from $P$ and $O^{dist}$ represents the distance between $S$ and $P$.}
\vspace*{\baselineskip}\endgroup

\subsection{Performance Computation}
\label{perfcomp}
We use F-score and perceptual quality (to resolve a few overlap boundary cases when we felt that the F-score is not discriminative) as the two main parameters to evaluate a model's performance. The F-score of crack pixels computed by choosing a threshold that results in the highest F-score over all the images considered for evaluation; has been used as the performance metric. It is also known as Overall Dataset Score (ODS) and the computation is adapted from here\footnote{\url{https://github.com/yhlleo/DeepSegmentor/blob/master/eval/prf_metrics.py}}.
It is to be noted that the performance metric (ODS) is chosen with respect to the task in consideration i.e. segmentation and a suitable performance metric can be chosen based on the required tasks like classification, object detection, etc.

\section{Experiments}
\label{experiments}

\subsection{Crack vs Non Crack Distinguishability}
\label{crackvnoncrack}
There is a significant difference in the visual representation of the crack and non-crack datasets, our goal is to show that the differences are captured by the selected pretrained models (namely SAM, CLIPSeg and ENet). To quantify these differences, we compute the $O^{dist}(S, P)$ (as discussed in section~\ref{distancecomp}) of each of the secondary datasets $S$ from the primary dataset $P$ using these models.

These pretrained models can capture the global contexts as they are trained on large datasets having different backgrounds. This suggests that we can use the distance metric to distinguish different datasets. However, these pretrained models are not trained to perform crack segmentation (for more details, see section~\ref{saminf} and~\ref{clipinf}). Therefore, to understand the relationship between the crack datasets, we selected three architectures namely DC, UNet$^{++}$ and ADSAM to train on $P$ and computed the distances of the secondary crack datasets from $P$ along with the performance on these secondary datasets.

DC is trained using scripts provided here\footnote{\url{https://github.com/qinnzou/DeepCrack}}, UNet$^{++}$ is initialized with an EfficientNet-B3 backbone~\ref{efficientnet} and  trained with a multi-gpu setup using segmentation models pytorch~\ref{smp} and pytorch lightning~\ref{plig} and ADSAM is trained using the "base" version of SAM~\ref{sam}. The training scripts for ADSAM can be found here\footnote{\url{https://github.com/chenyangzhu1/SAM-Adapter-PyTorch}}.

\subsection{Distance and Performance}

The distinguishability of images within each crack dataset for DC, UNet$^{++}$ and ADSAM is analyzed. The performance of these models is computed on the secondary crack datasets $S$ ($S_1, S_2, S_3$ and $S_4$). Within the same dataset, a set of images can be closer to $P$ whereas another set of images can be farther from $P$. It is expected that the model performance will be better on the images closer to $P$ as compared to those that are farther. To validate the same, we perform the following analysis.

\subsubsection{Intra Dataset Analysis}
\label{intrasplits}
Firstly, we compute the $I^{dist}(S, P)$. The images are then sorted in ascending order of distances $I^{dist}(S, P)$ and divided into two equal parts based on the distance. The F-score ($F_1$, $F_2$), mean distances ($\mu_1$, $\mu_2$) and standard deviations($\sigma_1$, $\sigma_2$) of each part are computed. The same computation is repeated by splitting the dataset into 3 equal parts to show the difference in performance between the closest and farthest images of $S$ from $P$. A plot~\ref{udidx} of sorted images vs distance from $P$ can be seen in the appendix.

\begin{align}
	\begin{split}
		\mu_1 = \frac{1}{\frac{m}{2}} \sum_{j=1}^{\frac{m}{2}}I_j^{dist}(S, P) \\
		\mu_2 = \frac{1}{\frac{m}{2}} \sum_{j=\frac{m}{2} + 1}^{m}I_j^{dist}(S, P)\\
		\sigma_1 = \frac{1}{m/2} \sum_{j=1}^{m/2} I_j^{dist}(S, P) - \mu_1\\
		\sigma_2 = \frac{1}{\frac{m}{2}} \sum_{j=\frac{m}{2} + 1}^{m} I_j^{dist}(S, P) - \mu_2\\
	\end{split}
\end{align}
\begingroup\vspace*{-\baselineskip}
\captionedequationset{($\mu_1$, $\mu_2$) and ($\sigma_1$, $\sigma_2$) are the means of the first half and second half of the sorted $I^{dist}(S, P)$. Computation can be performed for three parts similarly.}
\vspace*{\baselineskip}\endgroup

\subsection{Model Adaptation}
\label{qshot}
We perform an experiment by selecting $n$ (from P)+$q$ (from $S$) images ($n = \lvert P \rvert$ and $q \in \{1,3,7\}$) from each secondary dataset $S$. These datasets are used to train three models with different adaptations. The baseline models are herafter referred as M and the adapted models as M$_1$, M$_3$ and M$_7$ respectively. The F-score vs  $I^{dist}(S, P)$ plots of these adapted models are compared with the baseline models i.e. the models trained on just $P$. For plotting purposes, we perform a moving average with a window size of $\lvert S \rvert/10$ on the F-scores and  a min-max scaling on $I^{dist}(S, P)$. In min-max scaling, we scale a set of numbers x into a range of 0-1 by using the absolute minimum and maximum values of x and get $x_{scaled} = (x-x_{min})/(x_{max}-x_{min})$. The images were selected at a distance of 0.6-1 after the min-max scaling. 

\subsection{Model Understanding}
\label{fetbsds}
To understand the behaviour of the models on images different from the crack images, we perform inference using the models (DC, UNet$^{++}$ and ADSAM trained on $P$) on the images from the BSDS500~\ref{bsds500} dataset. The motivation is to observe the masks produced by these models and understand whether the model is actually predicting cracks or is predicting crack like features/edges from the images. If a model extracts crack like features from every image as cracks, it may give ambiguous distances.
We also compute the distances obtained from a variety of models on scene-centric scenarios to understand the model behaviour on diverse range of scenes(see details in section~\ref{scene}).

\section{Results}
\label{results}

\begin{figure}[htpb]
	\begin{subfigure}[htbp]{0.48\textwidth}
	\centering
	\includegraphics[width=\textwidth,height=0.12\textheight]{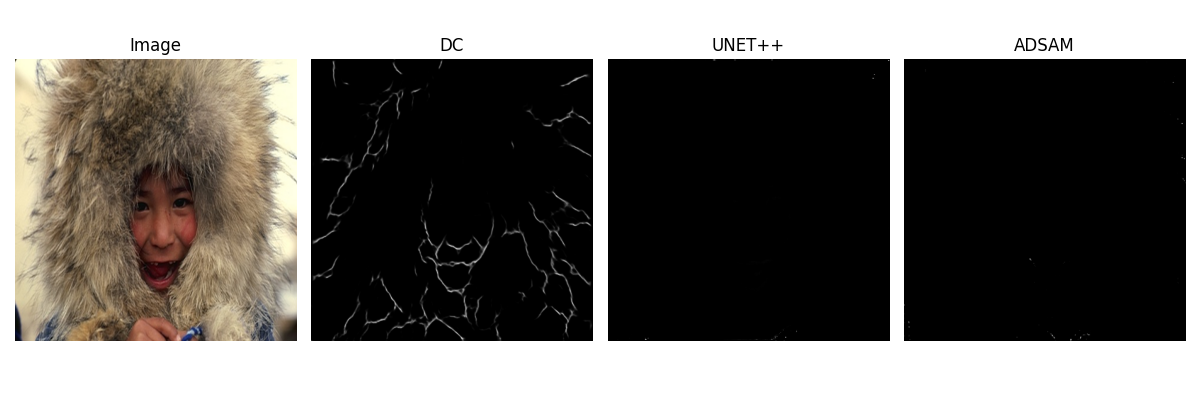}
	%\label{fig:12084}
	\end{subfigure}
	\begin{subfigure}[htbp]{0.48\textwidth}
	\centering
	\includegraphics[width=\linewidth,height=0.12\textheight]{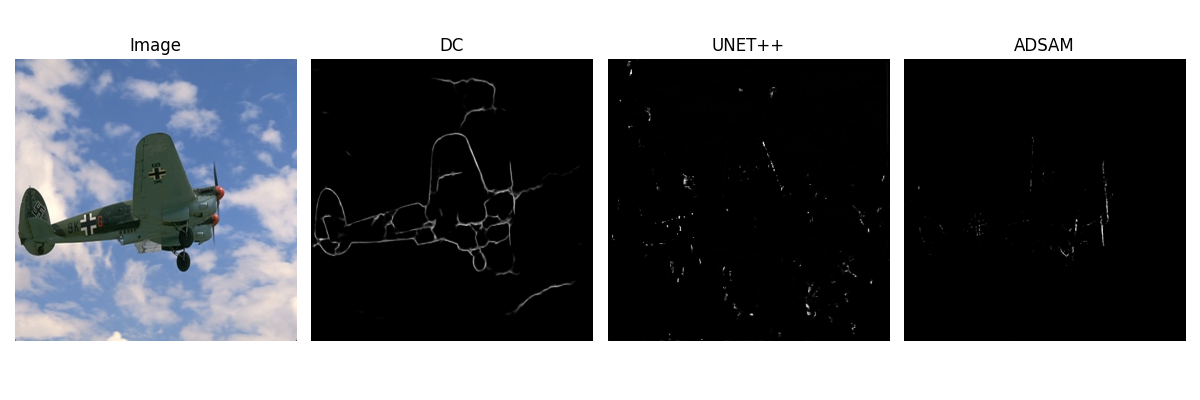}
	%\label{fig:3063}
	\end{subfigure}
	\begin{subfigure}[htbp]{0.48\textwidth}
		\centering
		\includegraphics[width=\linewidth,height=0.12\textheight]{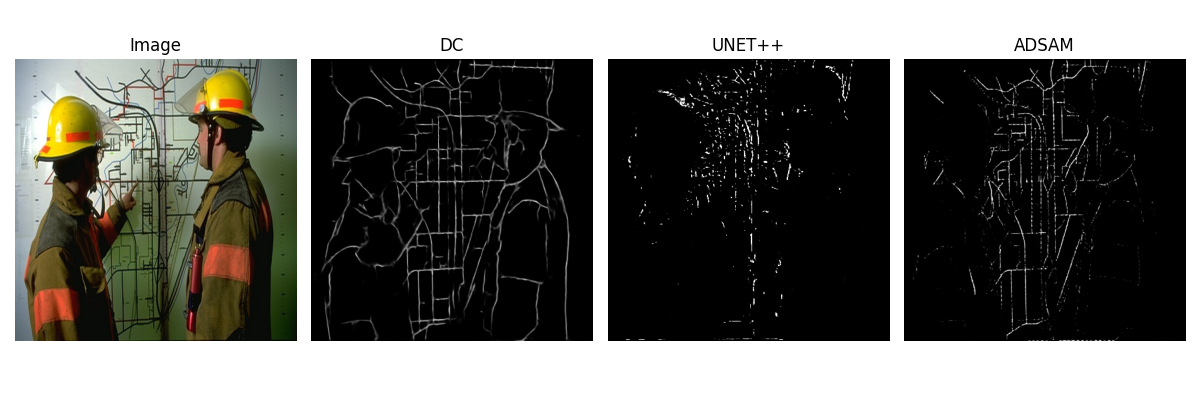}
		%\label{fig:23084}
	\end{subfigure}
	\begin{subfigure}[htbp]{0.48\textwidth}
		\centering
		\includegraphics[width=\linewidth,height=0.12\textheight]{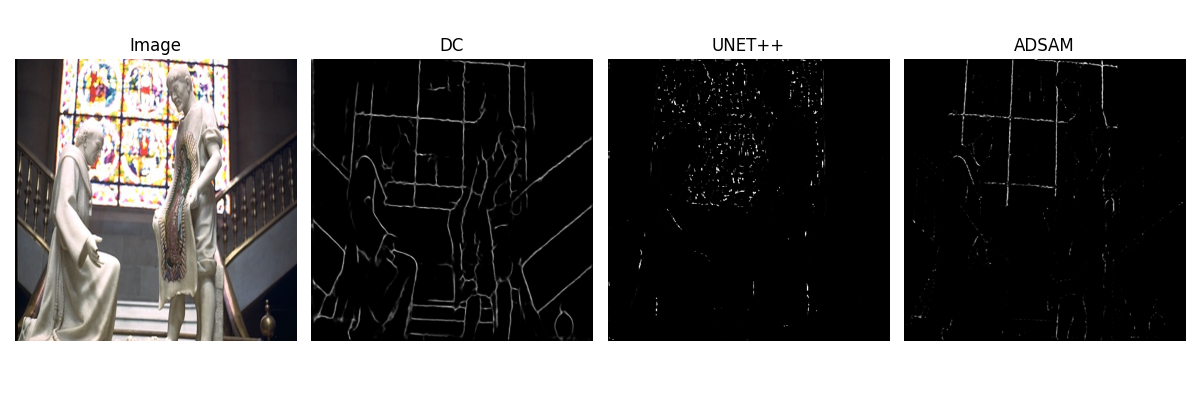}
		%\label{fig:24077}
	\end{subfigure}
	\caption{Shows (a) original image, (b) DC , (c) UNet$^{++}$ and (d) ADSAM outputs on the images from $S_6$. DC model focuses on specific features like edges.}
	\label{fig:udcadpred}
\end{figure}

\begin{figure}[h]
	\begin{subfigure}[t]{0.15\textwidth}
		\centering
		\includegraphics[width=\textwidth,height=0.12\textheight]{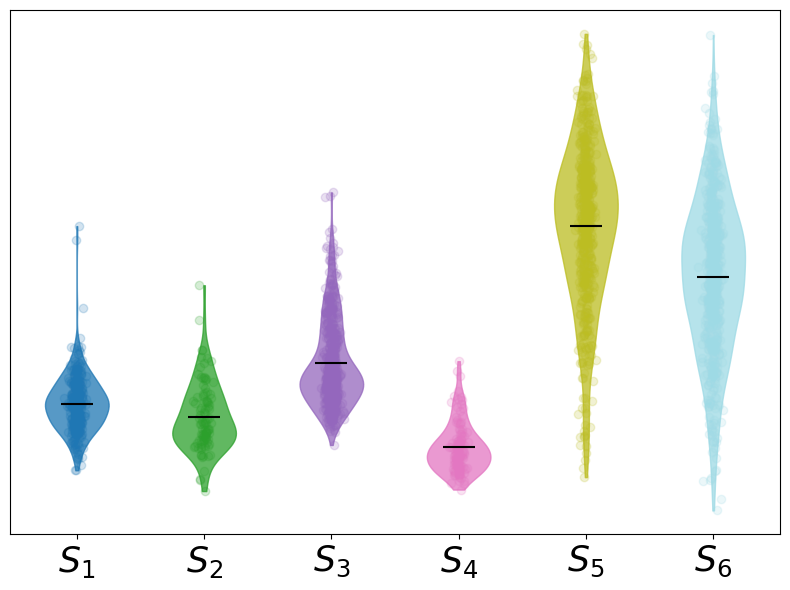}
		\caption{SAM}
		\label{fig:violinsam}
	\end{subfigure}
	\begin{subfigure}[t]{0.15\textwidth}
		\centering
		\includegraphics[width=\linewidth,height=0.12\textheight]{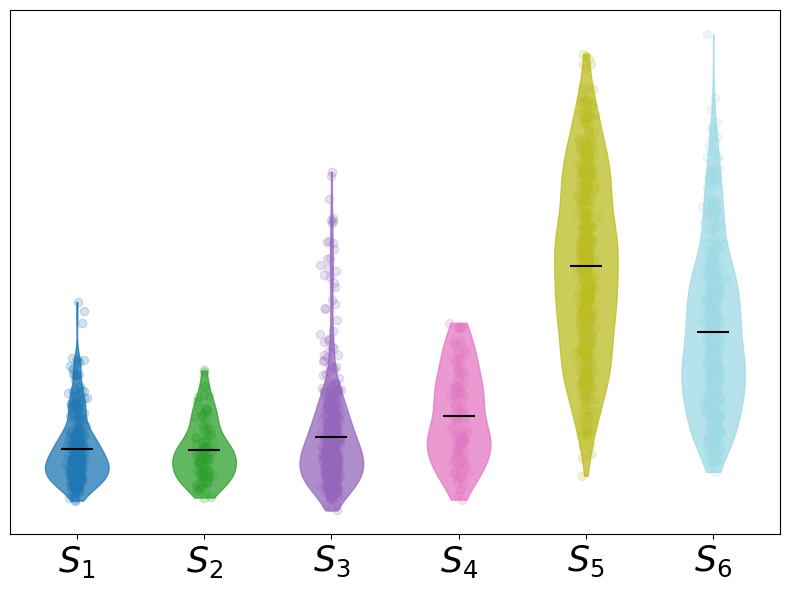}
		\caption{CLIPSeg}
		\label{fig:violinclip}
	\end{subfigure}
	\begin{subfigure}[t]{0.15\textwidth}
		\centering
		\includegraphics[width=\textwidth,height=0.12\textheight]{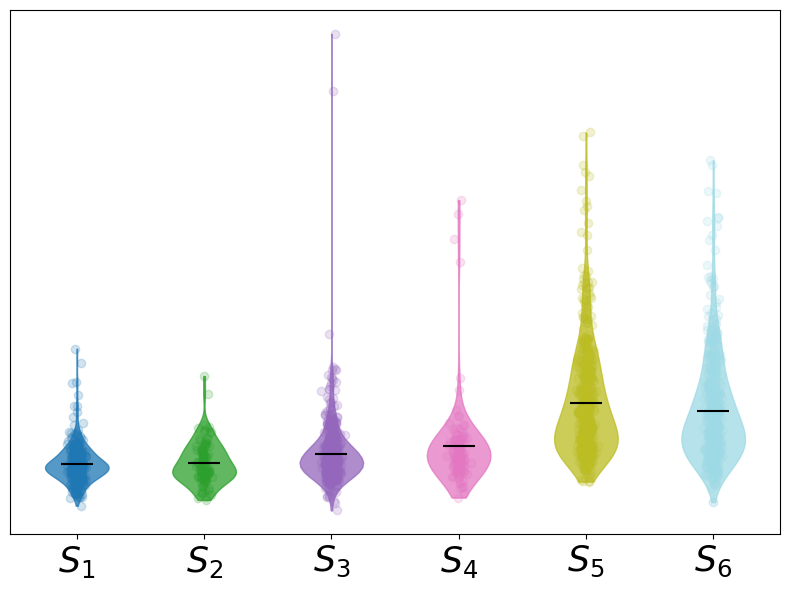}
		\caption{ENet}
		\label{fig:violinenet}
	\end{subfigure}
	\begin{subfigure}[t]{0.15\textwidth}
		\centering
		\includegraphics[width=\textwidth,height=0.12\textheight]{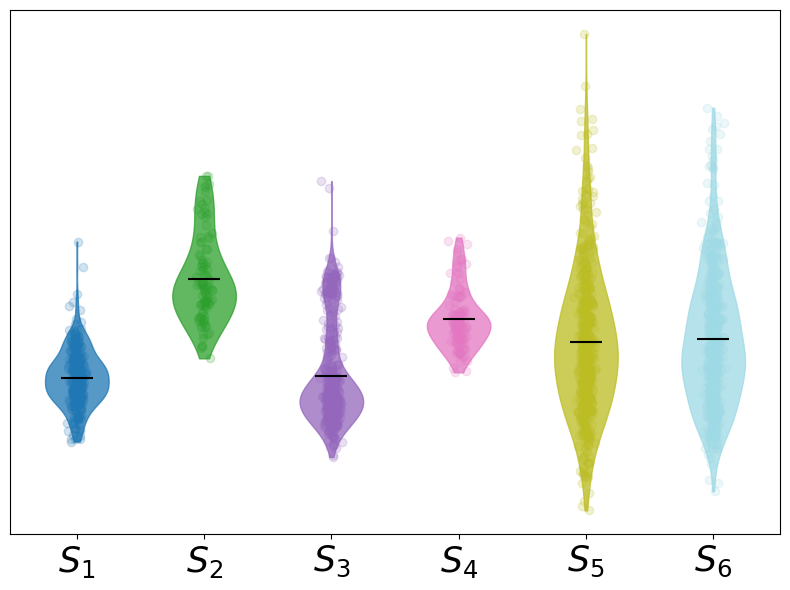}
		\caption{DC}
		\label{fig:violindc}
	\end{subfigure}
	\begin{subfigure}[t]{0.15\textwidth}
		\centering
		\includegraphics[width=\linewidth,height=0.12\textheight]{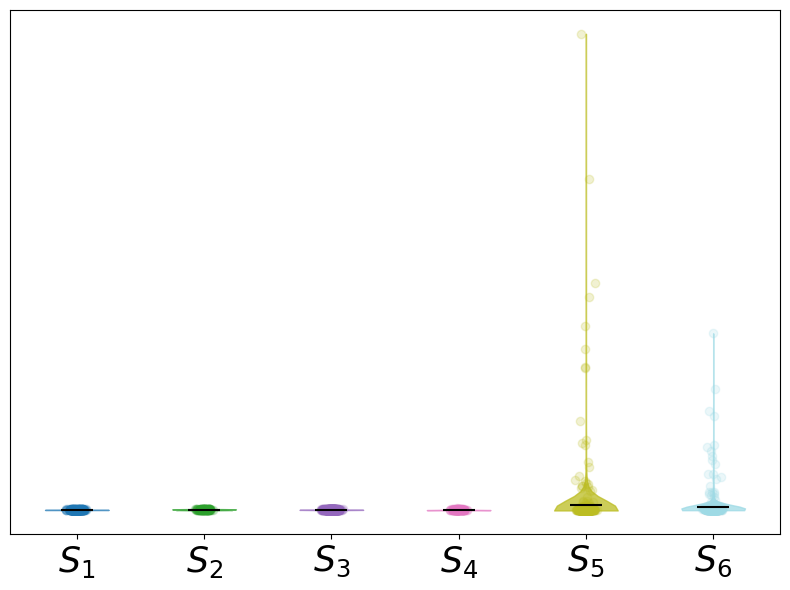}
		\caption{UNet$^{++}$}
		\label{fig:violinunet}
	\end{subfigure}
	\begin{subfigure}[t]{0.15\textwidth}
		\centering
		\includegraphics[width=\textwidth,height=0.12\textheight]{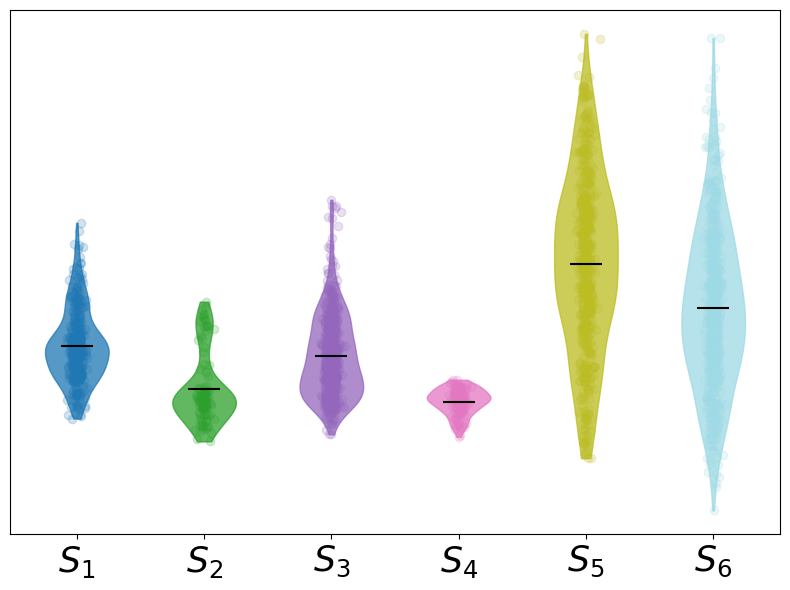}
		\caption{ADSAM}
		\label{fig:violinadsam}
	\end{subfigure}
	\caption{Shows the comparison of the violin plots (using ~\ref{plt}) of $I^{dist}$ distribution of all the secondary datasets  $S$ from $P$. The corresponding quantitative results can be seen in~\ref{distall}.}
	\label{fig:violin}
\end{figure}

% overall distance analysis
The table~\ref{distall} and figure~\ref{fig:violin} shows the distances computed from all the selected models (see details of the models in table~\ref{featurevector}). The non-crack datasets $S_5$ and $S_6$ are the farthest from $P$ for all the models except for DC where the model's focus is on the edges present in the images (see figure~\ref{fig:udcadpred}).

Since there is a tradeoff between the precision and recall, we can decide the kind of model required for the task in hand. The figure~\ref{fig:udcadpred} gives a good understanding on the behaviour of the models. In figure~\ref{fig:udcadpred}, the deepcrack model is more confident in detecting edge like structures compared to ADSAM and UNet$^{++}$. This suggests that ADSAM can be used for high precision and DC for high recall. 

% intra dataset analysis
The tables \ref{usplit} for UNet$^{++}$ , \ref{dsplit} for DC and \ref{adsplit} for ADSAM show the F-score and mean distance obtained by the models on each part of each secondary crack dataset (see captions of the tables for more details). The tables~\ref{usplit2} and~\ref{usplit3} show that the performance is also similar for the first and second half on $S_1$, $S_2$ and $S_3$ but the performance first half is significantly higher than the second half for $S_4$. Overall, there is a decreasing trend with distances for the UNet$^{++}$ model. Similarly, the other two models also follow a decreasing trend in the performance with the distances which can be seen in the tables~\ref{dsplit2},~\ref{dsplit3},~\ref{adsplit2} and~\ref{adsplit3}.

\begin{figure}[h]
	\centering
	\begin{subfigure}[h]{0.22\textwidth}
		\includegraphics[width=\linewidth,height=0.15\textheight]{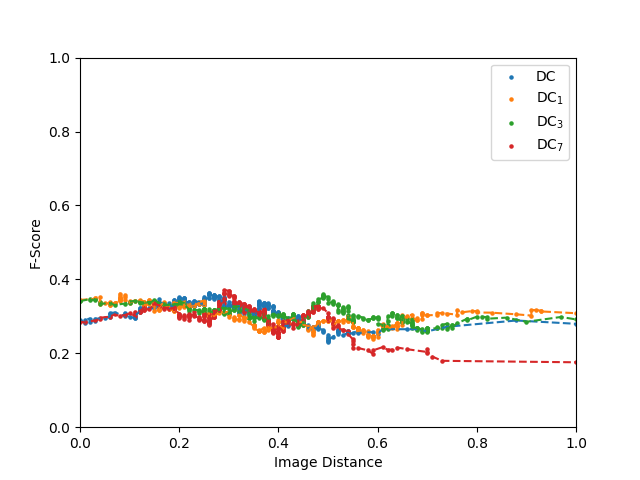}
		\caption{$S_1$}
		\label{fig:dc315}
	\end{subfigure}
	\begin{subfigure}[h]{0.22\textwidth}
		\includegraphics[width=\linewidth,height=0.15\textheight]{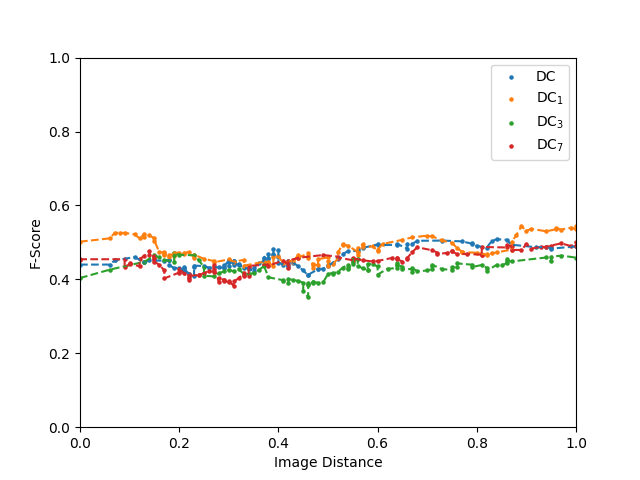}
		\caption{$S_2$}
		\label{fig:dcKWH100}
	\end{subfigure}
	\begin{subfigure}[h]{0.22\textwidth}
		\centering
		\includegraphics[width=\linewidth,height=0.15\textheight]{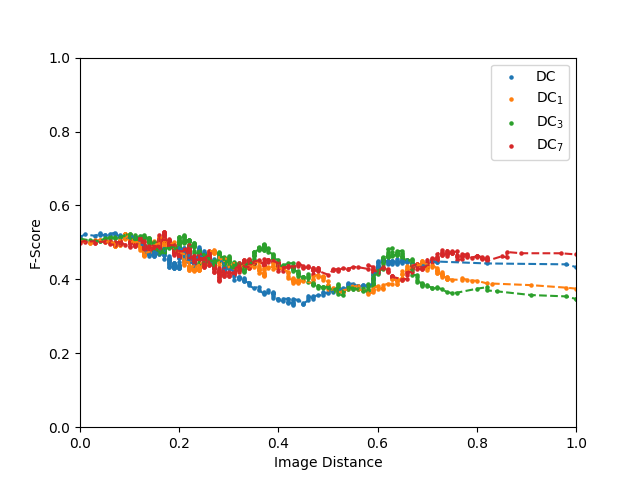}
		\caption{$S_3$}
		\label{fig:dcGAPs}
	\end{subfigure}
	\begin{subfigure}[h]{0.22\textwidth}
		\centering
		\includegraphics[width=\linewidth,height=0.15\textheight]{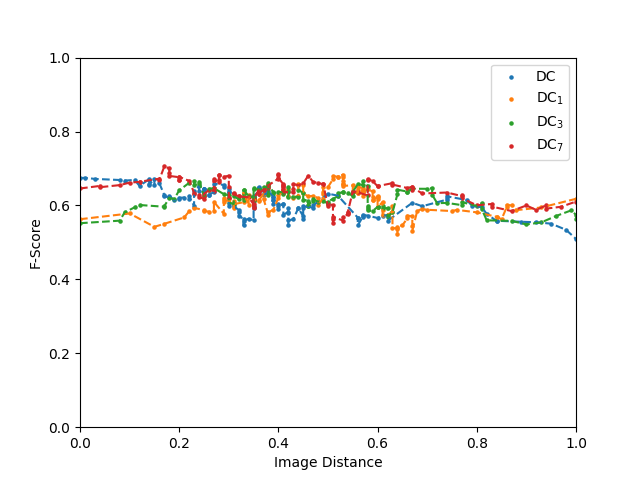}
		\caption{$S_4$}
		\label{fig:dcforest}
	\end{subfigure}	
	\caption{The figure shows the  F-score vs $I^{dist}(S, P)$ plot for $S$. The results are computed using the DC model which was trained on $P + n, n\in\{0,1,3,7\}$ where n images are selected from each $S$.}
	\label{fig:dc_if}
\end{figure}

\begin{figure}[h]
	\centering
	\begin{subfigure}[h]{0.22\textwidth}
		\includegraphics[width=\linewidth,height=0.15\textheight]{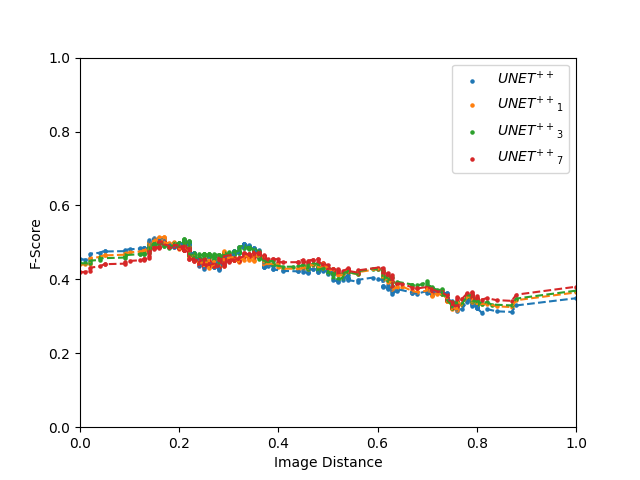}
		\caption{$S_1$}
		\label{fig:unet315}
	\end{subfigure}
	\begin{subfigure}[h]{0.22\textwidth}
		\includegraphics[width=\linewidth,height=0.15\textheight]{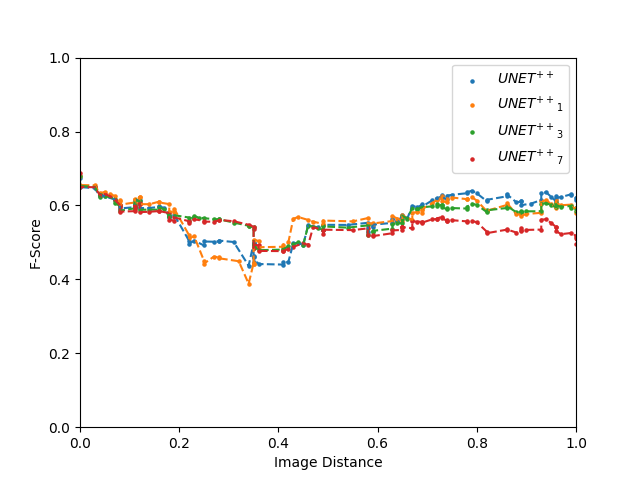}
		\caption{$S_2$}
		\label{fig:unetKWH100}
	\end{subfigure}
	\begin{subfigure}[h]{0.22\textwidth}
		\centering
		\includegraphics[width=\linewidth,height=0.15\textheight]{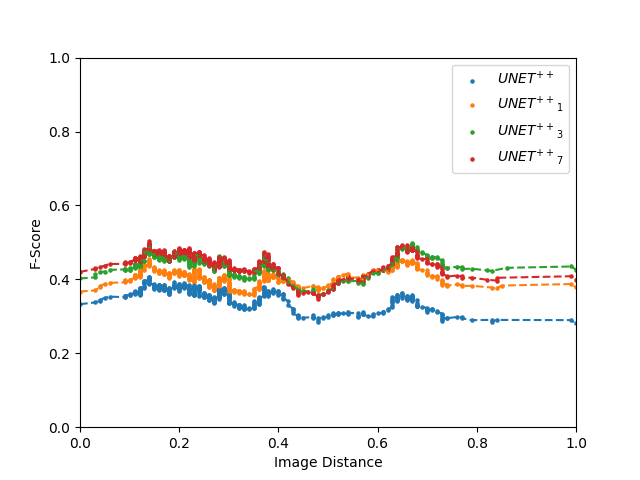}
		\caption{$S_3$}
		\label{fig:unetGAPs}
	\end{subfigure}
	\begin{subfigure}[h]{0.22\textwidth}
		\centering
		\includegraphics[width=\linewidth,height=0.15\textheight]{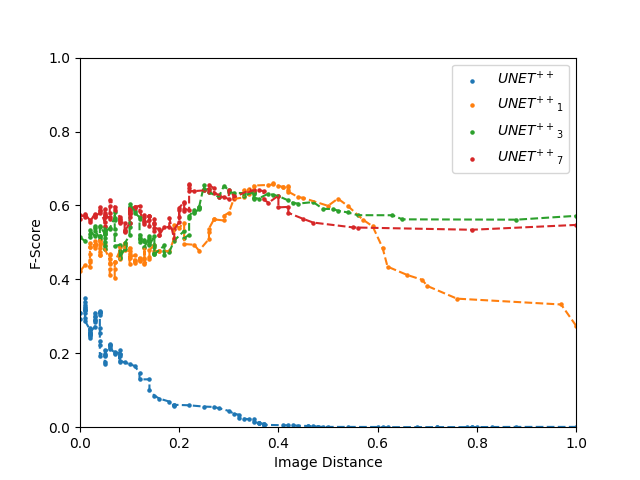}
		\caption{$S_4$}
		\label{fig:unetforest}
	\end{subfigure}	
	\caption{This figure shows the  F-score vs $I^{dist}(S, P)$ plot for $S$. The results are computed using UNet$^{++}$ model which was trained on $P + n, n\in\{0,1,3,7\}$ where n images are selected from each $S$.}
	\label{fig:unet_if}
\end{figure}

\begin{figure}[htbp!]
	\centering
	\begin{subfigure}[h]{0.22\textwidth}
		\includegraphics[width=\linewidth,height=0.15\textheight]{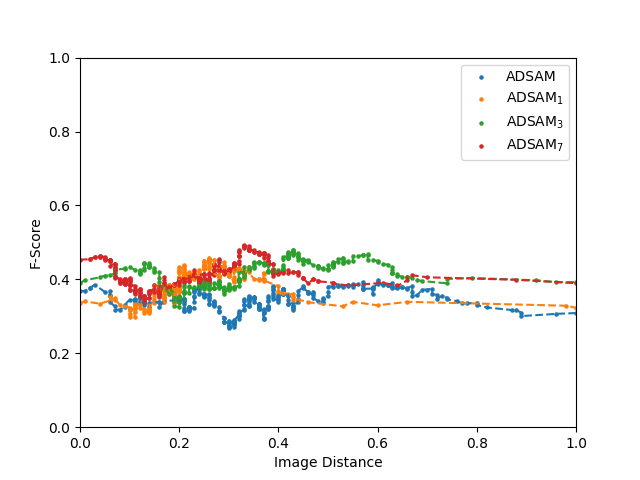}
		\caption{$S_1$}
		\label{fig:adsam315}
	\end{subfigure}
	\begin{subfigure}[h]{0.22\textwidth}
		\includegraphics[width=\linewidth,height=0.15\textheight]{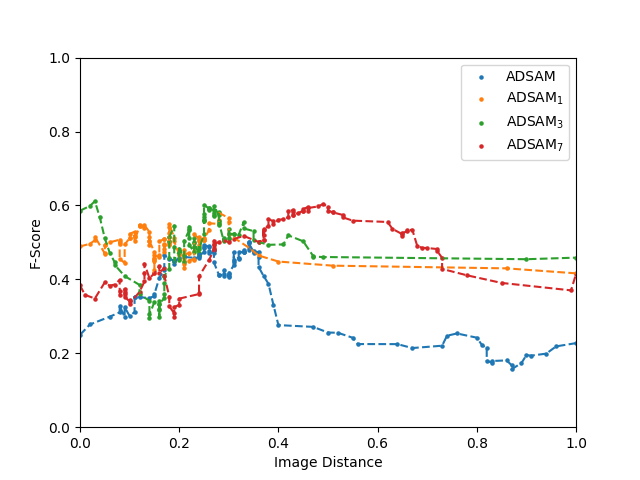}
		\caption{$S_2$}
		\label{fig:adsamKWH100}
	\end{subfigure}
	\begin{subfigure}[h]{0.22\textwidth}
		\centering
		\includegraphics[width=\linewidth,height=0.15\textheight]{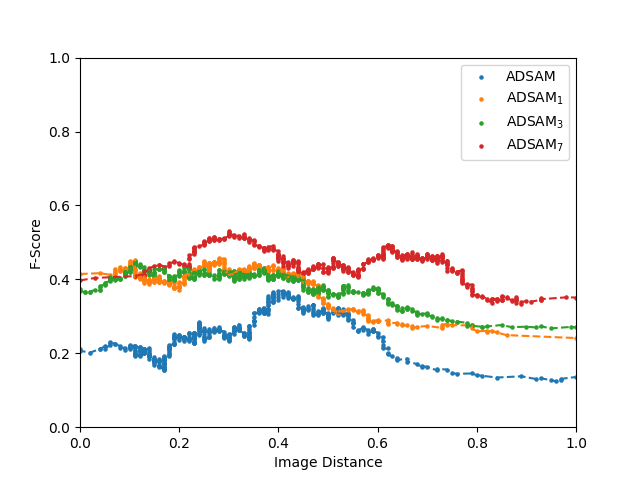}
		\caption{$S_3$}
		\label{fig:adsamGAPs}
	\end{subfigure}
	\begin{subfigure}[h]{0.22\textwidth}
		\centering
		\includegraphics[width=\linewidth,height=0.15\textheight]{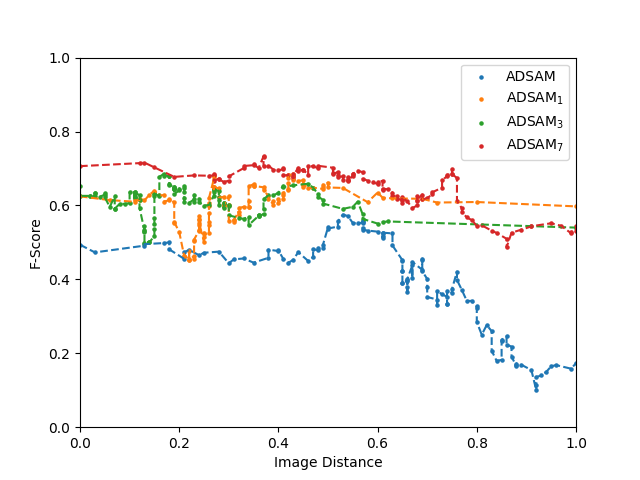}
		\caption{$S_4$}
		\label{fig:adsamforest}
	\end{subfigure}	
	\caption{This figure shows the F-score vs $I^{dist}(S, P)$ plot for $S$. The results are computed using the ADSAM model which was trained on $P + n, n\in\{0,1,3,7\}$ where n images are selected from each $S$.}
	\label{fig:adsam_if}
\end{figure}

\begin{figure}[htbp!]
	\centering
	\includegraphics[width=0.48\textwidth,height=0.15\textheight]{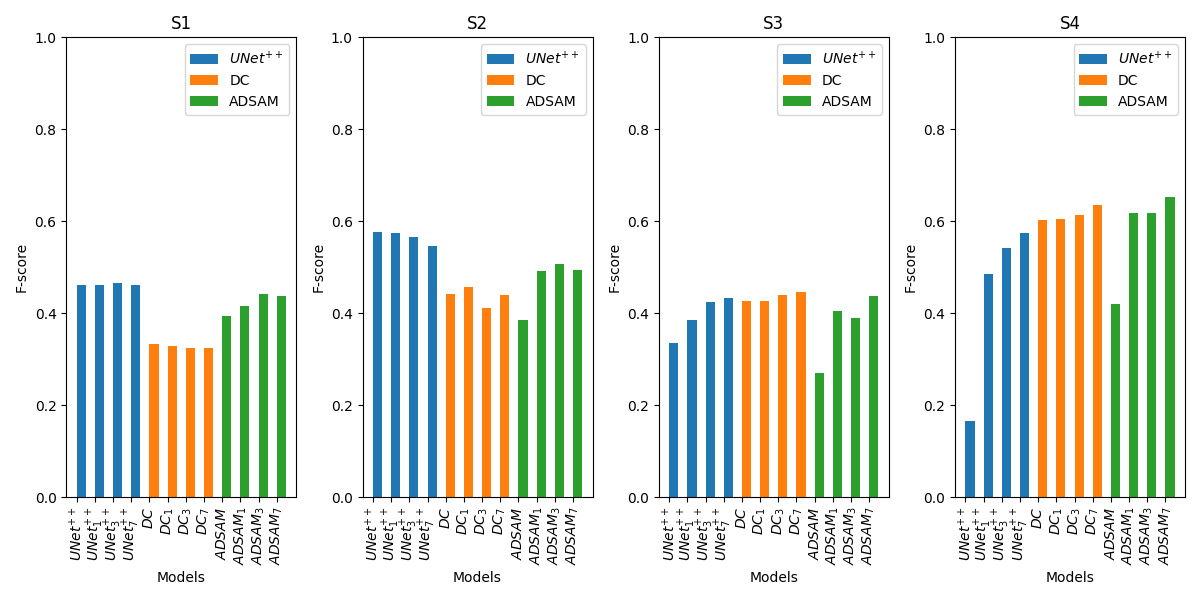}
	\caption{The figure shows the performance (F-score) of the M, M$_1$, M$_3$ and M$_7$ models on the secondary datasets, $S$.}
	\label{fig:fdist}
\end{figure}

% {1,3,7} images experiment
The figures~\ref{fig:dc_if},~\ref{fig:unet_if} and~\ref{fig:adsam_if} show the F-score vs $I^{dist}(S,P)$ comparison for the models UNet$^{++}$, DC and ADSAM respectively. These plots show the model performance when trained on $P$, $P$ + 1 image from each secondary, $P$ + 3 images from each secondary and $P$ + 7 images from each secondary dataset. After the training of the models, we can see a significant improvement in the performance across datasets and images, thereby increasing the generalization of the models (see captions and figure~\ref{fig:fdist} for additional details). 

\begin{table}[htb]
	\begin{tabular}{p{1.2cm}p{0.7cm}p{0.7cm}p{0.7cm}p{0.7cm}p{0.7cm}p{0.7cm}}
		\toprule
		& $S_1$ & $S_2$ & $S_3$ & $S_4$ & $S_5$ & $S_6$ \\
		\midrule
		SAM & 0.149& 0.145& 0.164& 0.134& \textbf{0.213} & \textbf{0.195}\\
		CLIPSeg & 0.135  &0.134  &0.142  &0.153  &\textbf{0.236}  &\textbf{0.199} \\
		ENet & 0.153& 0.153& 0.159& 0.163& \textbf{0.188} & \textbf{0.183} \\
		\midrule
		$DC$ & 0.153& \textbf{0.188} & 0.153& \textbf{0.174} & 0.166& 0.167\\
		UNet$^{++}$ & 0.056 & 0.069 & 0.063 & 0.054 & \textbf{0.465} & \textbf{0.293} \\
		ADSAM & 0.166 & 0.152 & 0.163 & 0.148 & \textbf{0.193} & \textbf{0.178} \\
		\bottomrule
	\end{tabular}
	\caption{The table shows the scaled $O^{dist}(S, P)$. The numbers in each cell are obtained after dividing the $O^{dist}(S, P)$ by the sum of the corresponding row. The top 2 farthest distances for each model from $P$ are shown in bold.}
	\label{distall}
\end{table}

\begin{table}
	\begin{subtable}[t]{0.4\textwidth}
		\begin{tabular}{lrr}
		\toprule
		Dataset & $F(\mu, \sigma)[0-50\%]$ & $F(\mu, \sigma)[50-100\%]$\\
		\midrule
		$S_1$ & 0.483(0.2, 0.06) & 0.444(0.54, 0.18) \\
		$S_2$ & 0.567(0.24, 0.15) & 0.606(0.79, 0.13) \\
		$S_3$ & 0.353(0.19, 0.06) & 0.316(0.48, 0.16) \\
		$S_4$ & 0.241(0.03, 0.02) & 0.061(0.4, 0.23) \\
		\bottomrule
	\end{tabular}
	\caption{The table shows the comparison of the performance of first 50\% and last 50\% images.}
	\label{usplit2}
	\end{subtable}
	\begin{subtable}[t]{0.4\textwidth}
		\begin{tabular}{lrr}
			\toprule
			Dataset & $F(\mu, \sigma)[0-33\%]$ & $F(\mu, \sigma)[67-100\%]$\\
			\midrule
			$S_1$ & 0.488(0.18, 0.05) & 0.415(0.64, 0.13) \\
			$S_2$ & 0.584(0.15, 0.09) & 0.622(0.85, 0.1) \\
			$S_3$ & 0.365(0.16, 0.04) & 0.314(0.56, 0.13) \\
			$S_4$ & 0.283(0.02, 0.01) & 0.021(0.5, 0.17) \\
			\bottomrule
		\end{tabular}
		\caption{The table shows the comparison of the performance of first 33\% and last 33\% images.}
		\label{usplit3}
	\end{subtable}
	\caption{The performance is computed using UNet$^{++}$ trained on $P$. Each cell in the tables represents the F-score(mean, standard deviation). See details in the section~\ref{intrasplits}.}
	\label{usplit}
\end{table}

\begin{table}
	\begin{subtable}[t]{0.4\textwidth}
		\begin{tabular}{lrr}
			\toprule
			Dataset & $F(\mu, \sigma)[0-50\%]$ & $F(\mu, \sigma)[50-100\%]$\\
			\midrule
			$S_1$ & 0.346(0.21, 0.08) & 0.318(0.42, 0.09) \\
			$S_2$ & 0.432(0.25, 0.09) & 0.464(0.62, 0.19) \\
			$S_3$ & 0.461(0.16, 0.04) & 0.391(0.42, 0.16) \\
			$S_4$ & 0.625(0.24, 0.09) & 0.58(0.56, 0.18) \\
			\bottomrule
		\end{tabular}
		\caption{The table shows the comparison of the performance of first 50\% and last 50\% images.}
		\label{dsplit2}
	\end{subtable}
	\begin{subtable}[t]{0.4\textwidth}
		\begin{tabular}{lrr}
			\toprule
			Dataset & $F(\mu, \sigma)[0-33\%]$ & $F(\mu, \sigma)[67-100\%]$\\
			\midrule
			$S_1$ & 0.34(0.17, 0.07) & 0.30(0.47, 0.1) \\
			$S_2$ & 0.43(0.2, 0.08) & 0.489(0.7, 0.16) \\
			$S_3$ & 0.472(0.14, 0.04) & 0.37(0.5, 0.14) \\
			$S_4$ & 0.638(0.19, 0.07) & 0.58(0.63, 0.17) \\
			\bottomrule
		\end{tabular}
		\caption{The table shows the comparison of the performance of first 33\% and last 33\% images.}
		\label{dsplit3}
	\end{subtable}
	\caption{The performance is computed using DC trained on $P$. Each cell in the tables represents the F-score(mean, standard deviation). See details in the section~\ref{intrasplits}.}
	\label{dsplit}
\end{table}

\begin{table}
	\begin{subtable}[t]{0.4\textwidth}
	\begin{tabular}{lrr}
		\toprule
		Dataset & $F(\mu, \sigma)[0-50\%]$ & $F(\mu, \sigma)[50-100\%]$\\
		\midrule
		$S_1$ & 0.388(0.23, 0.09) & 0.4(0.51, 0.14) \\
		$S_2$ & 0.392(0.2, 0.08) & 0.375(0.56, 0.24) \\
		$S_3$ & 0.249(0.19, 0.07) & 0.292(0.47, 0.14) \\
		$S_4$ & 0.471(0.44, 0.17) & 0.33(0.79, 0.1) \\
		\bottomrule
	\end{tabular}
	\caption{The table shows the comparison of the performance of first 50\% and last 50\% images.}
	\label{adsplit2}
	\end{subtable}
	\begin{subtable}[t]{0.4\textwidth}
		\begin{tabular}{lrr}
			\toprule
			Dataset & $F(\mu, \sigma)[0-33\%]$ & $F(\mu, \sigma)[67-100\%]$\\
			\midrule
			$S_1$ & 0.383(0.184, 0.07) & 0.405(0.584, 0.13) \\
			$S_2$ & 0.37(0.16, 0.07) & 0.29(0.654, 0.21) \\
			$S_3$ & 0.226(0.153, 0.05) & 0.293(0.533, 0.12) \\
			$S_4$ & 0.462(0.353, 0.15) & 0.291(0.842, 0.07) \\
			\bottomrule
		\end{tabular}
		\caption{The table shows the comparison of the performance of first 33\% and last 33\% images.}
		\label{adsplit3}
	\end{subtable}
	\caption{The performance is computed using ADSAM trained on $P$. Each cell in the tables represents the F-score(mean, standard deviation). See details in the section~\ref{intrasplits}.}
	\label{adsplit}
\end{table}

\begin{figure}[htpb]
	\begin{subfigure}[t]{0.48\textwidth}
		\centering
		\includegraphics[width=\textwidth,height=0.12\textheight]{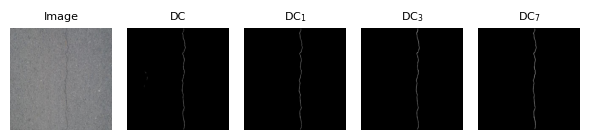}
		%\caption{DC}
		%\label{fig:dfor}
	\end{subfigure}
	\begin{subfigure}[t]{0.48\textwidth}
		\centering
		\includegraphics[width=\linewidth,height=0.12\textheight]{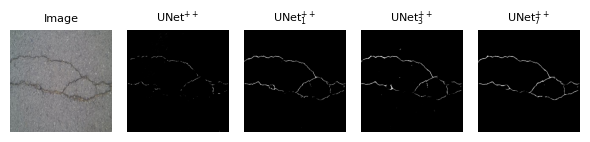}
		%\caption{UNet$^{++}$}
		%\label{fig:ufor}
	\end{subfigure}
	\begin{subfigure}[t]{0.48\textwidth}
		\centering
		\includegraphics[width=\textwidth,height=0.12\textheight]{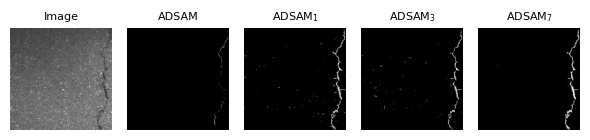}
		%\caption{ADSAM}
		%\label{fig:adfor}
	\end{subfigure}
	\caption{Shows original image, M, M$_1$, M$_3$ and M$_7$ outputs on different datasets. The input image for DC and UNet$^{++}$ is from $S_4$ and from $S_3$ for ADSAM. The titles of each figure (except the crack images from the dataset) show the name of model used for inference.}
	\label{fig:udcadforest}
\end{figure}

\begin{figure}[htpb]
	\begin{subfigure}[t]{0.48\textwidth}
		\centering
		\includegraphics[width=\textwidth,height=0.12\textheight]{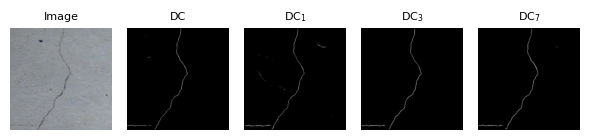}
		%\caption{DC}
		%\label{fig:deugen}
	\end{subfigure}
	\begin{subfigure}[t]{0.48\textwidth}
		\centering
		\includegraphics[width=\linewidth,height=0.12\textheight]{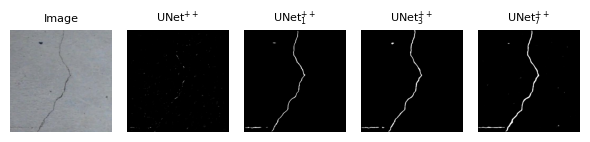}
		%\caption{UNet$^{++}$}
		%\label{fig:ueugen}
	\end{subfigure}
	\begin{subfigure}[t]{0.48\textwidth}
		\centering
		\includegraphics[width=\textwidth,height=0.12\textheight]{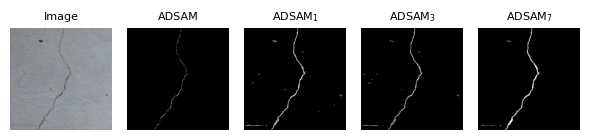}
		%\caption{ADSAM}
		%\label{fig:adeigen}
	\end{subfigure}
	\caption{Shows original image, Model, Model$_1$, Model$_3$ and Model$_7$ outputs on EUGEN MULLER \ref{kagglecrack} dataset. The titles of each figure (except the crack images from the dataset) show the name of model used for inference.}
	\label{fig:cfd}
\end{figure}

\section{Observations}
\label{observations}

The results in the figures~\ref{fig:dc_if},~\ref{fig:unet_if} and~\ref{fig:adsam_if} show that the performance of models trained on $P$ decreases as the distance increases. This gives a motivation to select a few images ($\{1,3,7\}$) from each of the secondary crack datasets and train the models by adding the selected images into the training data. The trained models seem to generalize on all secondary datasets which can be seen in the figures~\ref{fig:dc_if},~\ref{fig:unet_if} and~\ref{fig:adsam_if}. Fewer number of images are selected to reduce the labelling cost that can also improve the generalization of the models on different unseen datasets \ref{truedeep}.

The majority of $S_4$ images with an $I^{dist}(S_4, P)$ exceeding $0.4$ for the baseline UNet$++$ model and $0.5$ for the baseline $ADSAM$ model exhibit poor F-scores (as indicated by the blue line in figs.~\ref{fig:unet_if} (iv) and~\ref{fig:adsam_if} (iv)). As these models are adapted, there is a significant performance enhancement (orange, green and red lines in figs.~\ref{fig:unet_if} (iv),~\ref{fig:adsam_if} (iv)). Whereas, the performance for the DC model is almost same on all images in $S_4$ for the baseline and adapted models. As there is a tradeoff between crack and background prediction, some models produce high recall results and others, high precision. Therefore, some models require crack expression whereas others require suppression. The suppression analysis, discussed in the section~\ref{back} in the appendix, improves depth of our generalization study.

It's noticeable that as the distance from the training dataset increases, the performance of most models tends to decrease. This trend suggests that images closer to the training dataset generally exhibit better model performance, while those farther away tend to have lower performance. This decrease in performance can be attributed to the models being optimized for the training data, whereas unseen images might contain new patterns not covered during training. The F-score may not always be an accurate estimate of the performance and the ambiguity in some cases can be resolved using a different metric such as perceptual quality.

If the  performance plot vs distance is a near-horizontal line parallel to the x-axis, that indicates that the model is consistently performing well across the images (generalizes well). In this case, the model has already reached a satisfactory performance and adding new images from $S$ into the training set might not have a significant effect on the generalization of the model. However, the baseline performance can be further improved using additional set of images with a potential risk of overfitting. The amount of data required to achieve a near-horizontal line and the stability in performance metric across multiple unseen datasets are important factors that can help in the model selection.

The figures~\ref{fig:udcadforest} and~\ref{fig:cfd} show the qualitative improvement in performance of the adapted models over the baseline models i.e. the models only trained on $P$. The quality of predictions is in the order $M < M_1 < M_3 < M_7$ where M is the model used. The same trend can be observed in figure~\ref{fig:cfd} where the is improvement is shown on a completely blind dataset i.e. dataset that is not used for model training and adaptation. This shows that the overall adaptability of the models is increasing. The baseline DC model performs better than UNet$++$ and ADSAM but shows little improvement with adaptation whereas UNet$++$ and ADSAM show significant improvements with adaptation. Overall, the improvement in crack predictions using model adaptation suggests that there is an improvement in model generalization across multiple datasets. This can be also seen in the figures~\ref{fig:dc_if},~\ref{fig:unet_if} and~\ref{fig:adsam_if}, where the F-score vs $I_{dist}$ curve for the adapted models is near parallel to the x-axis. The improvement is not significant if the model is already performing consistently across unseen datasets which is the case for the DC model.

\section{Conclusion}
Our study analyzes multiple datasets to demonstrate that the performance of the models can be related to the proposed distance metrics (see figures~\ref{fig:dc_if},~\ref{fig:unet_if} and~\ref{fig:adsam_if}). This helps in improving an existing model with few images (improving generalization) and in turn, reducing the labeling and training cost while avoiding overfitting. We have shown that guiding the model with only a few images (see figure~\ref{fig:fdist}) from each of the secondary datasets can improve its performance significantly.

Our approach helps in explaining the behaviour of the models that can be seen in
figures~\ref{fig:udcadpred},~\ref{fig:violin},~\ref{fig:dc_if},~\ref{fig:unet_if} ,~\ref{fig:adsam_if} and~\ref{fig:cfd}. 

We also study the distances on scene-centric and person re-identification tasks (see details in sections~\ref{scene} and~\ref{personreid} in the Appendix).

\subsection{Future Scope}

The proposed approaches can be extended to explain model behaviour on other deep learning tasks by selecting our distance metrics (see section~\ref{distancecomp}) and a suitable performance metric (see section~\ref{perfcomp}) for that task. As the feature representations from the deep learning models express the input features, the applicability of the proposed distance metrics can be tested different data domains. 

The approaches can be further tested in real-world environments for examining the robustness and adaptability when faced with dynamic data distributions and generalization capabilities in the field. Research in this direction can have significant contributions in the applications that necessitate the selection of diverse datasets to train models capable of generalizing to new datasets.

Furthermore, theoretical aspects relating the distances to the model performance and generalization can also be studied.

\section{Acknowledgement}
We would like to thank GE Research for providing us with the resources and a conducive environment that made this research possible.

\printcredits

%\bibliography{}



\bibliographystyle{cas-model2-names}
\clearpage
\bibliography{00}
\begin{enumerate}[ {[}1{]} ]
	
	\item \label{ssim}
	Wang, Zhou, Alan C. Bovik, Hamid R. Sheikh, and Eero P. Simoncelli. "Image quality assessment: from error visibility to structural similarity." IEEE transactions on image processing 13.4 (2004): 600-612.
	
	\item \label{hog}
	Dalal, Navneet, and Bill Triggs. "Histograms of oriented gradients for human detection." IEEE computer society conference on computer vision and pattern recognition (CVPR'05). Vol. 1. Ieee, 2005.
	
	\item \label{siamese}
	Koch, Gregory, Richard Zemel, and Ruslan Salakhutdinov. "Siamese neural networks for one-shot image recognition." ICML deep learning workshop. Vol. 2. No. 1. 2015.
	
	\item \label{siamesematching}
	Melekhov, Iaroslav, Juho Kannala, and Esa Rahtu. "Siamese network features for image matching." 23rd international conference on pattern recognition (ICPR). IEEE, 2016.
	
	\item \label{otdd}
	Alvarez-Melis, David, and Nicolo Fusi. "Geometric dataset distances via optimal transport." Advances in Neural Information Processing Systems 33 (2020): 21428-21439.
	
	\item \label{dmlgen}
	Huai, Mengdi, et al. "Deep Metric Learning: The Generalization Analysis and an Adaptive Algorithm." IJCAI. 2019.
	
	\item \label{dmlnet}
	Cen, Jun, et al. "Deep metric learning for open world semantic segmentation." Proceedings of the IEEE/CVF International Conference on Computer Vision. 2021.
	
	\item \label{visgenunets}
	Rajagopal, Abhejit, et al. "Understanding and Visualizing Generalization in UNets." Medical Imaging with Deep Learning. PMLR, 2021.
	
	\item \label{simperf}
	Gevaert, Caroline M., and Mariana Belgiu. "Assessing the generalization capability of deep learning networks for aerial image classification using landscape metrics." International Journal of Applied Earth Observation and Geoinformation 114 (2022): 103054.
	
	\item \label{eqperf}
	Petersen, E., Holm, S., Ganz, M., \& Feragen, A. (2023). The path toward equal performance in medical machine learning. Patterns, 4(7).
	
	\item \label{ugenvis}
	Huang, W. R., Emam, Z., Goldblum, M., Fowl, L., Terry, J. K., Huang, F., and Goldstein, T. (2020). Understanding generalization through visualizations.
	
	\item \label{sam}
	Kirillov, Alexander, et al. "Segment anything." arXiv preprint arXiv:2304.02643 (2023).
	
	\item \label{adsam}
	Chen, Tianrun, et al. "SAM Fails to Segment Anything?--SAM-Adapter: Adapting SAM in Underperformed Scenes: Camouflage, Shadow, and More." arXiv preprint arXiv:2304.09148 (2023).
	
	\item \label{pascalvoc}
	Everingham, Mark, and John Winn. "The PASCAL visual object classes challenge 2012 (VOC2012) results. 2012
	
	\item \label{bsds500}
	Arbelaez, Pablo, et al. "Contour detection and hierarchical image segmentation." IEEE transactions on pattern analysis and machine intelligence 33.5 (2010): 898-916.
	
	\item \label{deepcrack}
	Zou, Qin, Zheng Zhang, Qingquan Li, Xianbiao Qi, Qian Wang, and Song Wang. "Deepcrack: Learning hierarchical convolutional features for crack detection." IEEE Transactions on Image Processing 28.3 (2018): 1498-1512.
	
	\item \label{gaps}
	Eisenbach, Markus, Ronny Stricker, Daniel Seichter, Karl Amende, Klaus Debes, Maximilian Sesselmann, Dirk Ebersbach, Ulrike Stoeckert, and Horst-Michael Gross. "How to get pavement distress detection ready for deep learning? A systematic approach." In 2017 international joint conference on neural networks (IJCNN), pp. 2039-2047. IEEE, 2017.
	
	\item \label{kagglecrack}
	'Kaggle Crack Segmentation Dataset',\\ https://www.kaggle.com/datasets/lakshaymiddha/crack-segmentation-dataset, 
	accessed 2020
	
	\item \label{clipseg}
	Lüddecke, Timo, and Alexander S. Ecker. "Prompt-based multi-modal image segmentation." arXiv preprint arXiv:2112.10003 (2021).
	
	\item \label{efficientnet}
	Tan, Mingxing, and Quoc Le. "Efficientnet: Rethinking model scaling for convolutional neural networks." International conference on machine learning. PMLR, 2019.
	
	\item \label{imagenet}
	Deng, Jia, et al. "Imagenet: A large-scale hierarchical image database." IEEE conference on computer vision and pattern recognition. Ieee, 2009.
	
	\item \label{unet++}
	Zhou, Zongwei, et al. "Unet++: A nested u-net architecture for medical image segmentation." Deep Learning in Medical Image Analysis and Multimodal Learning for Clinical Decision Support: 4th International Workshop, DLMIA 2018.

	\item \label{clip}
	Radford, Alec, Jong Wook Kim, Chris Hallacy, Aditya Ramesh, Gabriel Goh, Sandhini Agarwal, Girish Sastry,  Amanda Askell, Pamela Mishkin, Jack Clark, Gretchen Krueger, Ilya Sutskever. "Learning transferable visual models from natural language supervision." International conference on machine learning. PMLR, 2021.
	
	\item \label{mae}
	He, Kaiming, et al. "Masked autoencoders are scalable vision learners." Proceedings of the IEEE/CVF conference on computer vision and pattern recognition. 2022.
	
	\item \label{scikit}
	Pedregosa, Fabian, et al. "Scikit-learn: Machine learning in Python." the Journal of machine Learning research 12 (2011): 2825-2830.
	
	\item \label{smp}
	Iakubovskii, P. "Segmentation Models Pytorch." GitHub (2019).
	
	\item \label{plig}
	Falcon, William A. "Pytorch lightning." GitHub 3 (2019).
	
	\item \label{plt}
	Hunter, J. D. Matplotlib: A 2D graphics environment. Computing in science \& engineering, 9(03), 90-95. (2007).
	
	\item \label{truedeep}
	Pandey, Ramkrishna, and Akshit Achara. "TrueDeep: A systematic approach of crack detection with less data." Expert Systems with Applications (2023): 122785.
	
	\item \label{psam}
	Zhang, Renrui, et al. "Personalize segment anything model with one shot." arXiv preprint arXiv:2305.03048 (2023).
	
	\item \label{scenerec}
	Xie, Lin, et al. "Scene recognition: A comprehensive survey." Pattern Recognition 102 (2020): 107205.
	
	\item \label{sunpaper}
	Xiao, Jianxiong, et al. "Sun database: Large-scale scene recognition from abbey to zoo." 2010 IEEE computer society conference on computer vision and pattern recognition. IEEE, 2010.
	
	\item \label{dinov2}
	Oquab, Maxime, et al. "Dinov2: Learning robust visual features without supervision." arXiv preprint arXiv:2304.07193 (2023).
	
	\item \label{convnextv2}
	Woo, Sanghyun, et al. "Convnext v2: Co-designing and scaling convnets with masked autoencoders." Proceedings of the IEEE/CVF Conference on Computer Vision and Pattern Recognition. 2023.
	
	\item \label{swinv2}
	Liu, Ze, et al. "Swin transformer v2: Scaling up capacity and resolution." Proceedings of the IEEE/CVF conference on computer vision and pattern recognition. 2022.
	
	\item \label{places}
	Zhou, Bolei, et al. "Places: A 10 million image database for scene recognition." IEEE transactions on pattern analysis and machine intelligence 40.6 (2017): 1452-1464.
	
	\item \label{reidsurvey}
	Ming, Zhangqiang, et al. "Deep learning-based person re-identification methods: A survey and outlook of recent works." Image and Vision Computing 119 (2022): 104394.
	
	\item \label{market}
	Zheng, Liang, et al. "Scalable person re-identification: A benchmark." Proceedings of the IEEE international conference on computer vision. 2015.
	
	\item \label{cuhk03}
	Li, Wei, et al. "Deepreid: Deep filter pairing neural network for person re-identification." Proceedings of the IEEE conference on computer vision and pattern recognition. 2014.
	
	\item \label{unreal}
	Zhang, Tianyu, et al. "Unrealperson: An adaptive pipeline towards costless person re-identification." Proceedings of the IEEE/CVF Conference on Computer Vision and Pattern Recognition. 2021.
	
	\item \label{synperson}
	Xiang, Suncheng, et al. "Rethinking illumination for person re-identification: A unified view." Proceedings of the IEEE/CVF Conference on Computer Vision and Pattern Recognition. 2022.
	
	\item \label{reid}
	Xiang, Suncheng, et al. "Less is more: Learning from synthetic data with fine-grained attributes for person re-identification." ACM Transactions on Multimedia Computing, Communications and Applications 19.5s (2023): 1-20.
	
\end{enumerate}

\clearpage

% Use the style of numbering in square brackets.
% If nothing is used, default style will be taken.
%\begin{enumerate}[a)]
%\item 
%\item 
%\item 
%\end{enumerate}  

%\begin{itemize}
%\item 
%\item 
%\item 
%\end{itemize}  

%\begin{description}
%\item[]
%\item[] 
%\item[] 
%\end{description}  

% Uncomment and use as the case may be
%\begin{theorem} 
%\end{theorem}

% Uncomment and use as the case may be
%\begin{lemma} 
%\end{lemma}

%\clearpage
\appendix
\section{Distances}
\subsection{Effect of the Number of Principal Components on Distances}
\label{dimselection}
The selection of the number of principal components can be divided into three cases.
\begin{enumerate}
	\item \textbf{Case 1:}
	If the number of principal components selected for the distance computation is less than 15, it is observed that there is a significant difference in the distances with change in the number of principal components.
	\item \textbf{Case 2:}
	If the number of principal components selected for the distance computation is between 15 and 25, there is very less difference in the distances with the change in number of principal components.
	\item \textbf{Case 3:}
	If the number of principal components selected for the distance computation is greater than 25, there is negligible difference in the distances with the change in number of principal components.
\end{enumerate}
The table~\ref{pcdiff} shows the decrease in difference of distances between the subsequent dimensions.

\begin{table}[htbp!]
	\begin{tabular}{lrrrrrr}
		\toprule
		PC& $S_1$ & $S_2$ & $S_3$ & $S_4$ & $S_5$ & $S_6$ \\
		\midrule
		5        &0.078     &0.097  &0.087   &0.073   &0.453    &0.214\\
		10       &0.068     &0.084  &0.077   &0.065   &0.486    &0.220\\
		15       &0.063     &0.077  &0.071   &0.061   &0.491    &0.237\\
		20       &0.057     &0.071  &0.064   &0.055   &0.466    &0.287\\
		25       &0.056     &0.069  &0.063   &0.054   &0.465    &0.293\\
		\bottomrule
	\end{tabular}
	\caption{The table shows the values of $O^{dist}(S,P)$ computed by selecting varying number of principal components(PC). The model used here is UNet$^{++}$ trained on $P$. Each value in the table is divided by the sum of its row.}
	\label{pcdiff}
\end{table}
\subsection{Distance Distributions}
\begin{figure}[htbp!]
	\centering
	\begin{subfigure}[htbp]{0.2\textwidth}
		\includegraphics[width=\linewidth,height=0.12\textheight]{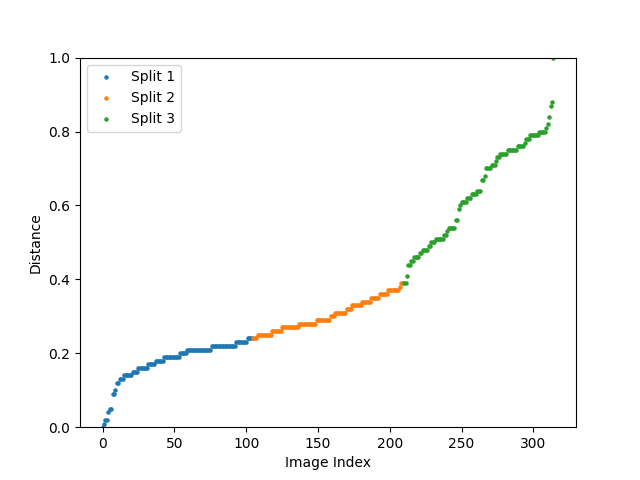}
		\caption{$S_1$}
		\label{fig:ud315}
	\end{subfigure}
	\begin{subfigure}[htbp]{0.2\textwidth}
		\includegraphics[width=\linewidth,height=0.12\textheight]{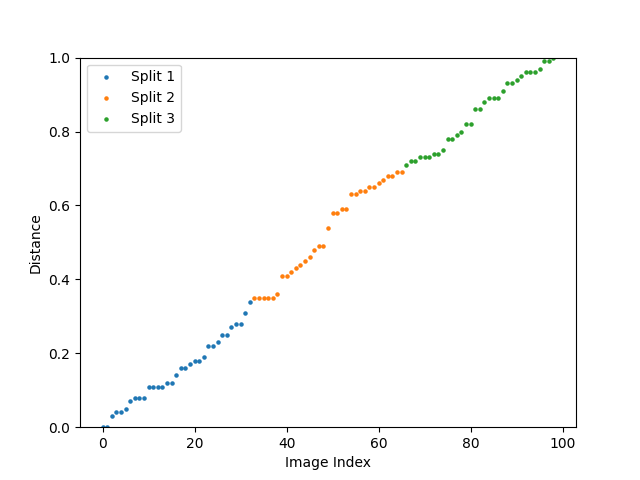}
		\caption{$S_2$}
		\label{fig:udKWH100}
	\end{subfigure}
	\begin{subfigure}[htbp]{0.2\textwidth}
		\centering
		\includegraphics[width=\linewidth,height=0.12\textheight]{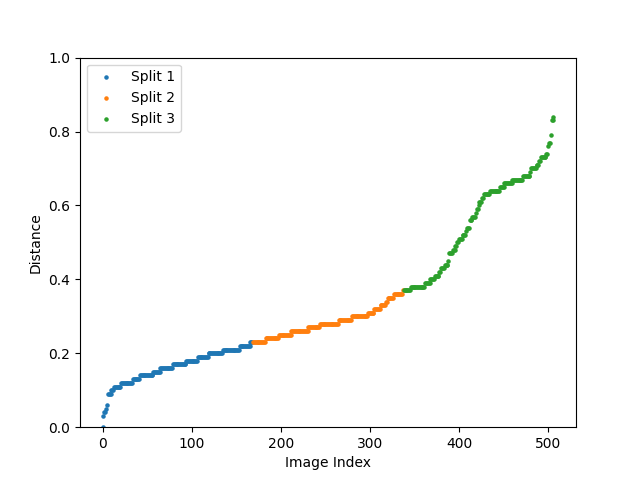}
		\caption{$S_3$}
		\label{fig:udGAPs}
	\end{subfigure}
	\begin{subfigure}[htbp]{0.2\textwidth}
		\centering
		\includegraphics[width=\linewidth,height=0.12\textheight]{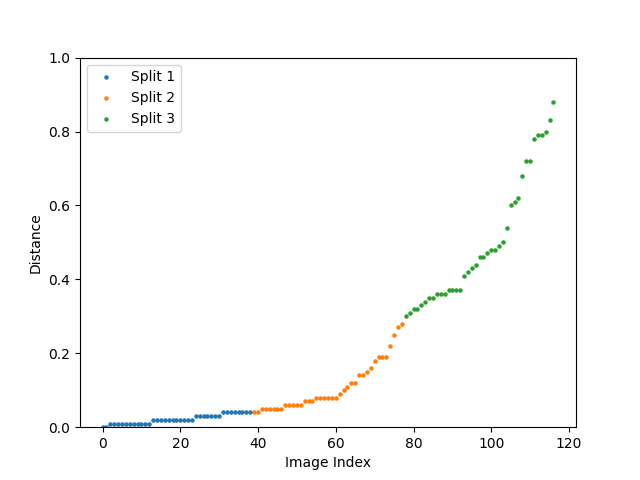}
		\caption{$S_4$}
		\label{fig:udforest}
	\end{subfigure}	
	\caption{The figure shows the distribution of the distances of images in each secondary crack dataset $S$ from the primary dataset $P$. The distances are computed using UNet$^{++}$ trained on $P$.}
	\label{udidx}
\end{figure}

It can be seen in figure~\ref{udidx} that there is a steep increase in the distance of the images in the third split for $S_4$ for which there is a significant difference in the F-score of the first 33\% and last 33\% images. A consistent distribution can be seen for $S_2$ where all splits had similar F-scores. In general, the consistency and inconsistency in the distance distribution of the datasets is reflected in the performance (F-score). See table~\ref{usplit3} for more details.

\section{Are Cracks an Important Feature for Models?}
\subsection{SAM}
\label{saminf}
\begin{figure}[htpb!]
	\centering
	\begin{subfigure}[t]{0.2\textwidth}
		\includegraphics[width=\linewidth,height=0.12\textheight]{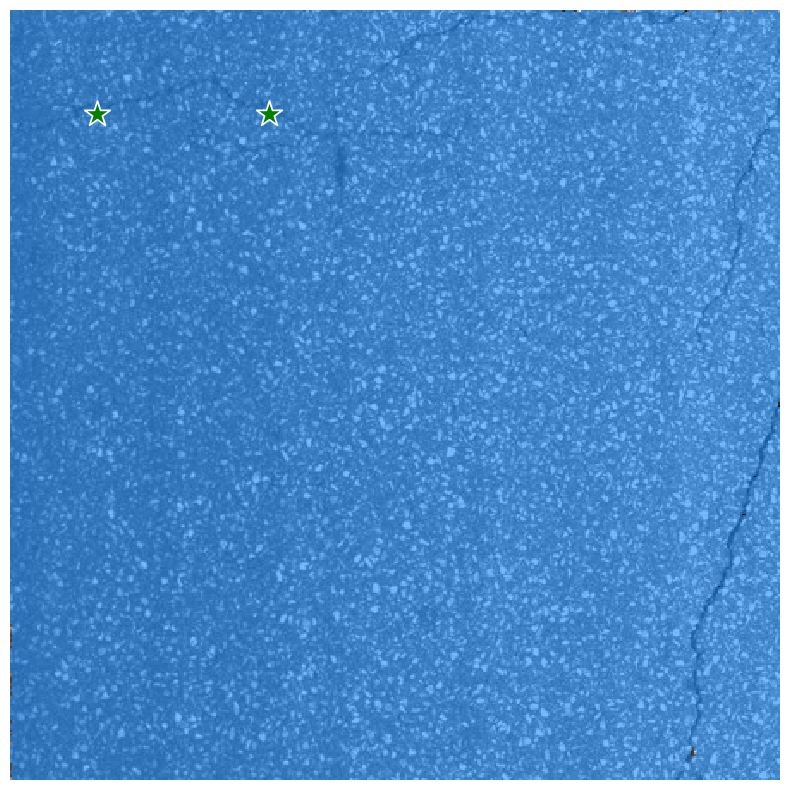}
		\caption{Prompted with two foreground points.}
		\label{fig:points}
	\end{subfigure}
	\begin{subfigure}[t]{0.2\textwidth}
		\includegraphics[width=\linewidth,height=0.12\textheight]{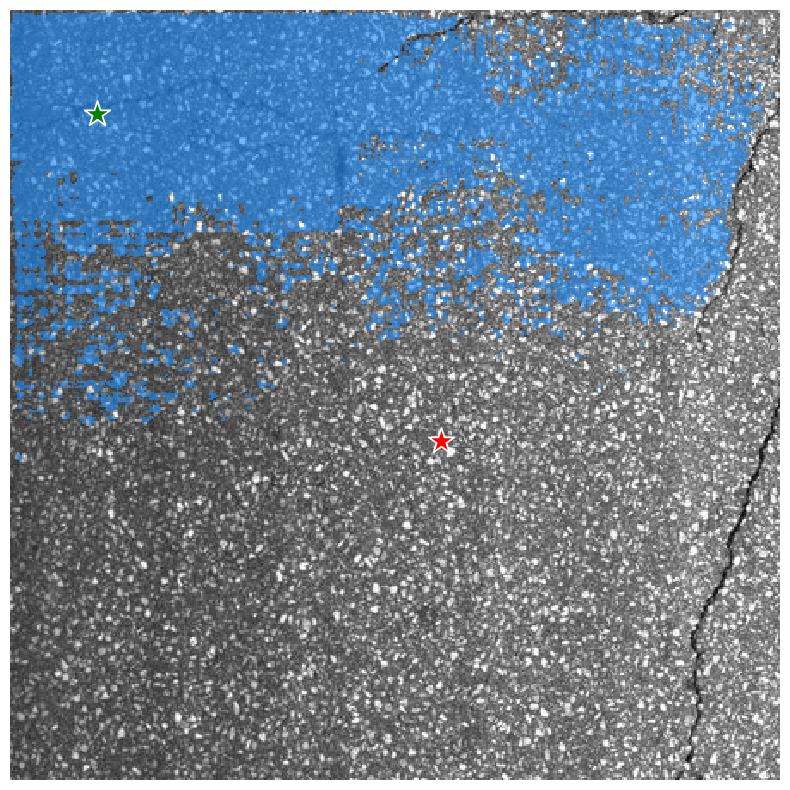}
		\caption{Prompted with a background point and foreground point. }
		\label{fig:ipoints}
	\end{subfigure}
	\begin{subfigure}[t]{0.2\textwidth}
		\centering
		\includegraphics[width=\linewidth,height=0.12\textheight]{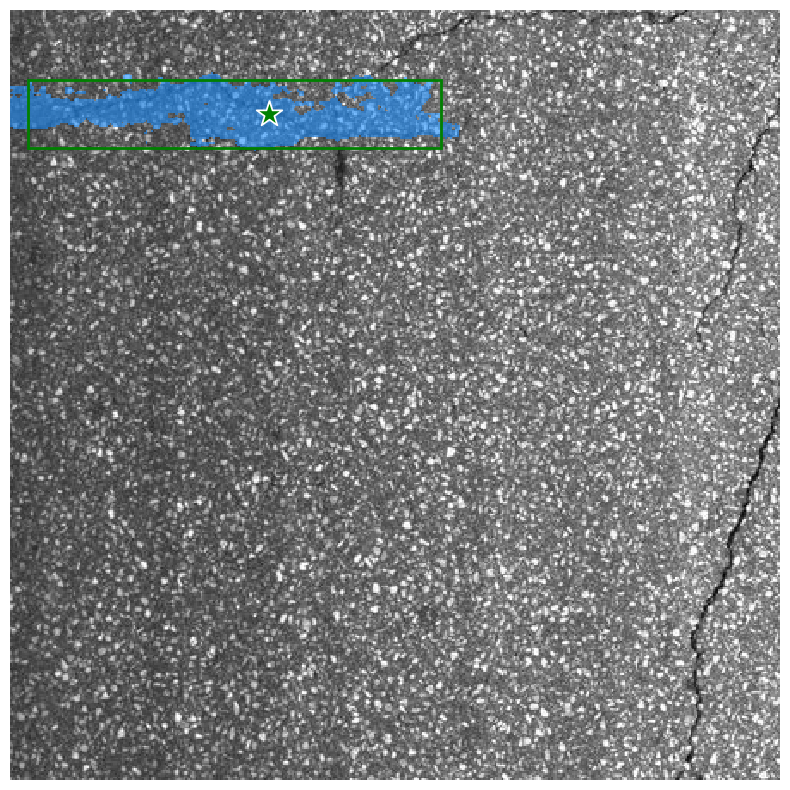}
		\caption{Prompted with a box and foreground point in it.}
		\label{fig:box}
	\end{subfigure}
	\begin{subfigure}[t]{0.2\textwidth}
		\centering
		\includegraphics[width=\linewidth,height=0.12\textheight]{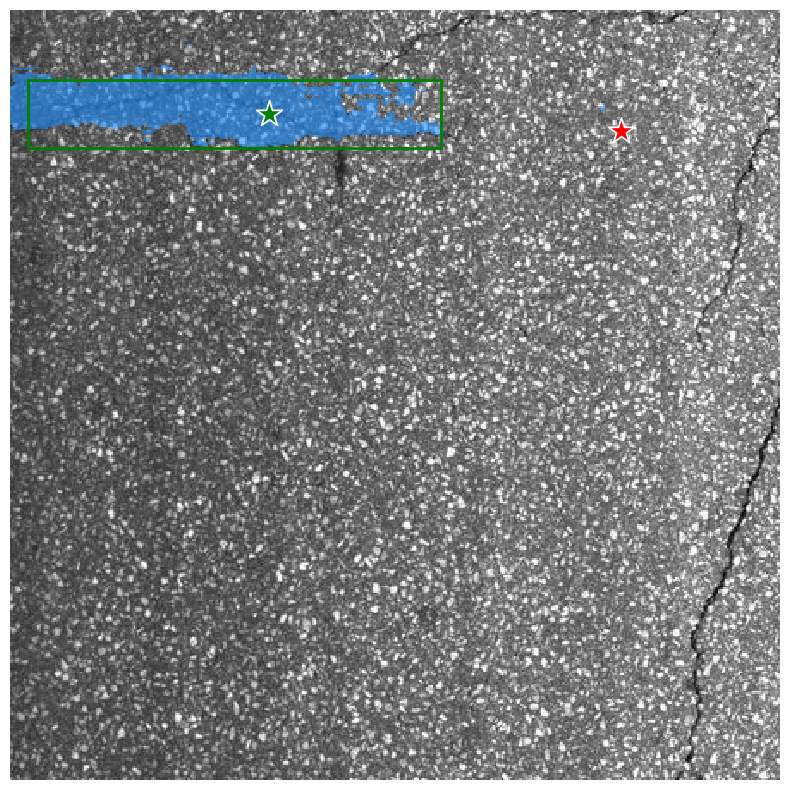}
		\caption{Prompted with a box containing a foreground point and a background point outside the box.}
		\label{fig:ibox}
	\end{subfigure}
	\caption{The figure shows the predicted masks from the SAM model by inputting an image from $S_1$. It can be observed that there is a focus on the surface (sometimes on the intricate details on the texture). However, cracks are not segmented here. Green points represent the foreground (crack) label prompt and red points represent the background label points.}
	\label{samcrack}
\end{figure}

The SAM can take points or boxes as prompts to produce output masks for input images. We give crack images as inputs to SAM along with prompt points and bounding boxes as shown in figure~\ref{samcrack} and observe that the output masks are surfaces or the intricacies of the surface rather than cracks. 
It is also difficult to provide prompt points at exact crack pixels across multiple images using an automated process. It is to be noted that there was no tuning performed on SAM with crack data in this case. We have performed one-shot finetuning of SAM on $P$ (see \ref{psam} for more details). There is no significant improvement in the crack detection. Finally, we have trained an adapted version of SAM (ADSAM \ref{adsam}) to perform crack detection. ADSAM detects cracks from secondary crack datasets $S$ similar to the other crack detection models.

\subsection{CLIPSeg}
\label{clipinf}
\begin{figure*}[htbp!]
	\centering
	\begin{subfigure}[htbp!]{0.75\textwidth}
		\centering
		\includegraphics[width=\linewidth,height=0.11\textheight]{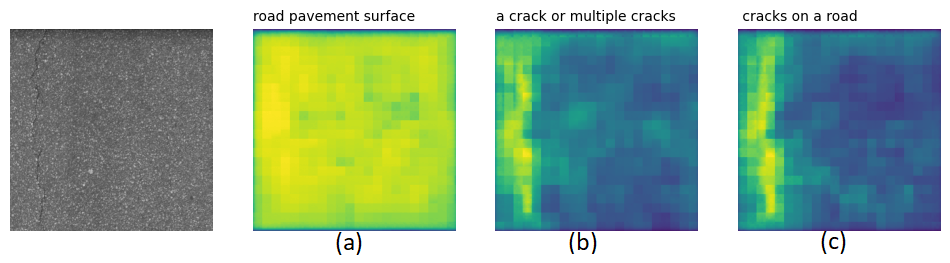}
		\caption{}
		\label{fig:textdif}
	\end{subfigure}
	\begin{subfigure}[htbp!]{0.75\textwidth}
		\centering
		\includegraphics[width=\linewidth,height=0.11\textheight]{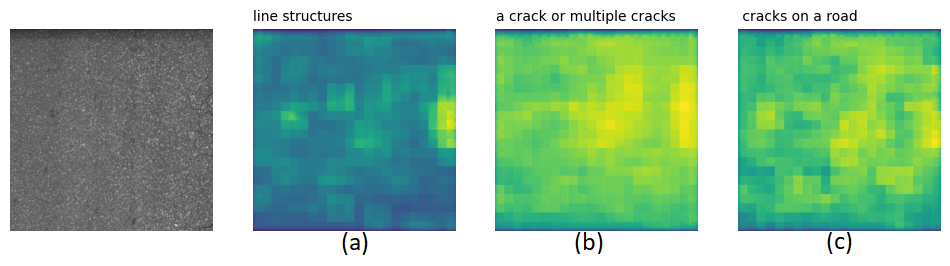}
		\caption{}
		\label{fig:clippred1}
	\end{subfigure}
	\begin{subfigure}[htbp!]{0.75\textwidth}
		\centering
		\includegraphics[width=\linewidth,height=0.11\textheight]{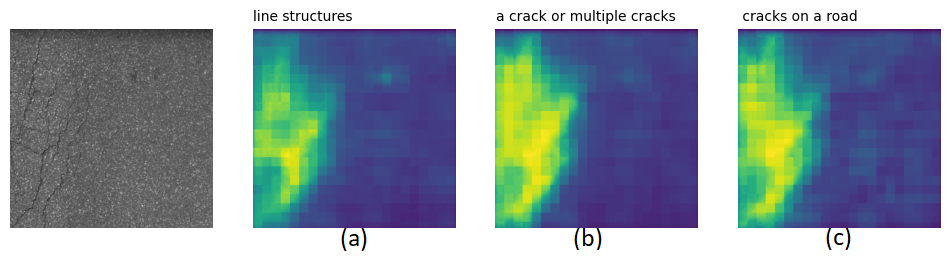}
		\caption{}
		\label{fig:clippred2}
	\end{subfigure}
	%\begin{subfigure}[htbp]{0.48\textwidth}
	%	\centering
	%	\includegraphics[width=\linewidth,height=0.11\textheight]{clip_am3}
	%	\caption{}
	%	\label{fig:clippred3}
	%\end{subfigure}
	\caption{Shows (a) original image, (b) mask produced with prompt 1, (c) mask produced with prompt 2, (d) mask produced with prompt 3. The prompts can be seen in titles of the respective masks.}
\end{figure*}

We give multiple text prompts along with input crack images to the CLIPSeg model for zero-shot inference ~\ref{clipseg}. The output probability maps highlight the crack regions but also capture background surfaces along the boundaries and other regions. Figure~\ref{fig:textdif} shows the effect of input text prompts where the text prompt given for (a) is 'road pavement surface' and the model segments the surface whereas the prompts given for (b) and (c) are 'a crack or multiple cracks' and 'cracks on a road' where the model segments the crack regions. Figures~\ref{fig:clippred1} and ~\ref{fig:clippred2} are not detecting the detecting the cracks at all or the boundaries are fuzzy. This suggests that the model needs to be trained for crack detection to perform similar to crack detection models. Finetuning of CLIPSeg requires labeled phrase region pairs that are not available for crack datasets.

\section{Suppressing Noise}
\label{back}

We conduct experiments to improve background prediction in our models. We introduce randomly selected 3 and 7 ambiguous non-crack surface images with blank masks from dataset \ref{kagglecrack}. These images are added to the original training data for UNet$_7^{++}$, DC$_7$, and ADSAM$_7$ models. Subsequently, these updated datasets train UNet$_{B3}^{++}$, DC$_{B3}$, and ADSAM$_{B3}$ (with $7\times 4$ + 3 additional images) and UNet$_{B7}^{++}$, DC$_{B7}$, and ADSAM$_{B7}$ (with $7\times 4$ + 7 additional images), referred to as M$_{B3}$ and M$_{B7}$ hereafter.

The goal is to reduce the background noise predicted by the models while maintaining the crack detection performance. 

\begin{figure}[pos=htpb!,align=\centering]
	\begin{subfigure}[t]{0.48\textwidth}
		\raggedright
		\includegraphics[width=\textwidth,height=0.12\textheight]{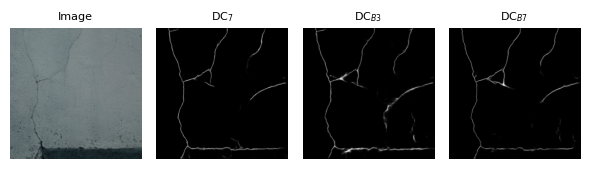}
		%\caption{DC}
		%\label{fig:deugen}
	\end{subfigure}
	\begin{subfigure}[t]{0.48\textwidth}
		\raggedright
		\includegraphics[width=\linewidth,height=0.12\textheight]{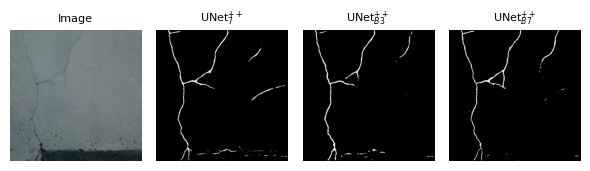}
		%\caption{UNet$^{++}$}
		%\label{fig:ueugen}
	\end{subfigure}
	\begin{subfigure}[t]{0.48\textwidth}
		\raggedright
		\includegraphics[width=\textwidth,height=0.12\textheight]{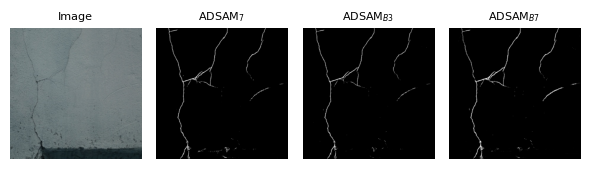}
		%\caption{ADSAM}
		%\label{fig:adeigen}
	\end{subfigure}
	\caption{Shows original image, M$_7$, M$_{B3}$ and M$_{B7}$ outputs on Rissbilder \ref{kagglecrack} dataset. The titles of each figure (except the crack images from the dataset) show the name of model used for inference.}
	\label{fig:nonriss}
\end{figure}

While models M$_1$, M$_3$, and M$_7$ enhance crack detection over baseline models (M), they lack training examples involving ambiguous structures resembling cracks but not actual cracks. Including such examples reduces noise in predictions, as evident in blind dataset inferences shown in the figure~\ref{fig:nonriss}. However, in some cases, both noise and actual cracks can be removed, leading to lower performance. This highlights the need for diverse examples in model adaptation, attainable by selecting a few images.

\begin{figure}[h]
	\centering
	\begin{subfigure}[h]{0.22\textwidth}
		\includegraphics[width=\linewidth,height=0.15\textheight]{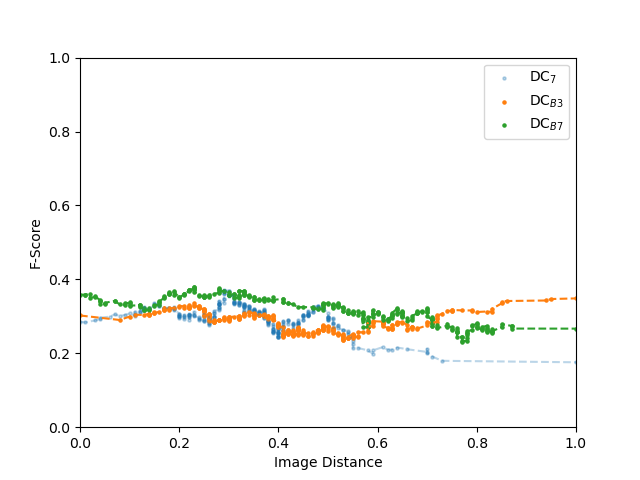}
		\caption{$S_1$}
		\label{fig:nondc315}
	\end{subfigure}
	\begin{subfigure}[h]{0.22\textwidth}
		\includegraphics[width=\linewidth,height=0.15\textheight]{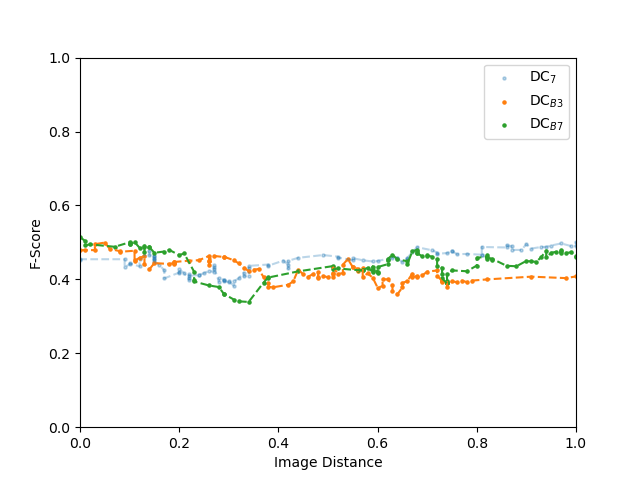}
		\caption{$S_2$}
		\label{fig:nondcKWH100}
	\end{subfigure}
	\begin{subfigure}[h]{0.22\textwidth}
		\centering
		\includegraphics[width=\linewidth,height=0.15\textheight]{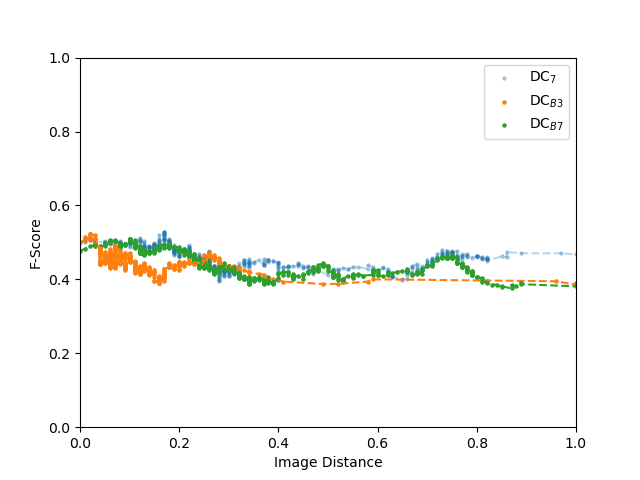}
		\caption{$S_3$}
		\label{fig:nondcGAPs}
	\end{subfigure}
	\begin{subfigure}[h]{0.22\textwidth}
		\centering
		\includegraphics[width=\linewidth,height=0.15\textheight]{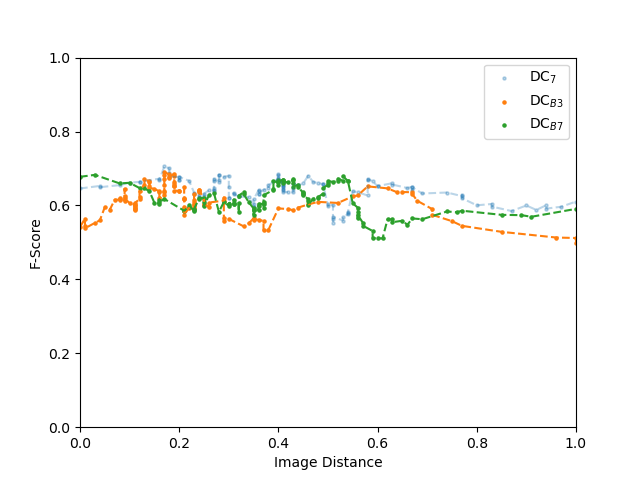}
		\caption{$S_4$}
		\label{fig:nondcforest}
	\end{subfigure}	
	\caption{The figure shows the  F-score vs $I^{dist}(S, P)$ plot for $S$. The results are computed using the DC$_7$, DC$_{B3}$ and DC$_{B7}$ models.}
	\label{fig:nondc_if}
\end{figure}

\begin{figure}[h]
	\centering
	\begin{subfigure}[h]{0.22\textwidth}
		\includegraphics[width=\linewidth,height=0.15\textheight]{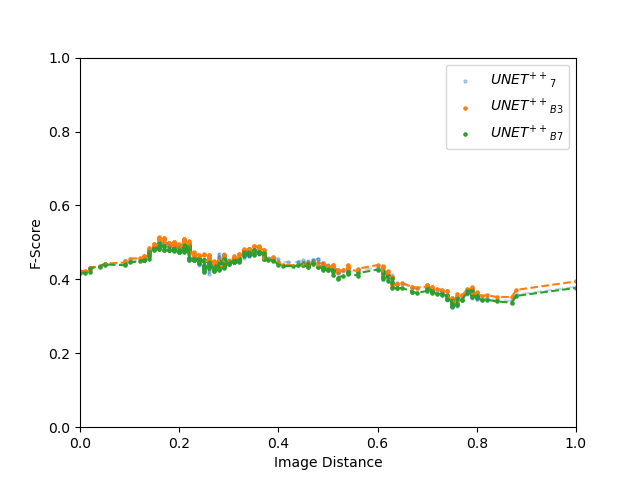}
		\caption{$S_1$}
		\label{fig:nonunet315}
	\end{subfigure}
	\begin{subfigure}[h]{0.22\textwidth}
		\includegraphics[width=\linewidth,height=0.15\textheight]{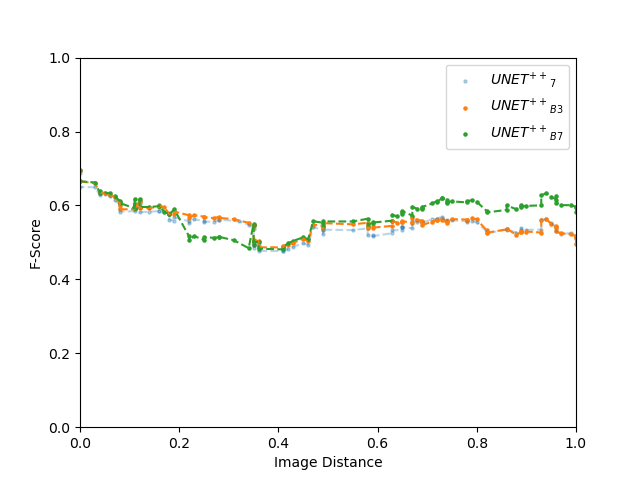}
		\caption{$S_2$}
		\label{fig:nonunetKWH100}
	\end{subfigure}
	\begin{subfigure}[h]{0.22\textwidth}
		\centering
		\includegraphics[width=\linewidth,height=0.15\textheight]{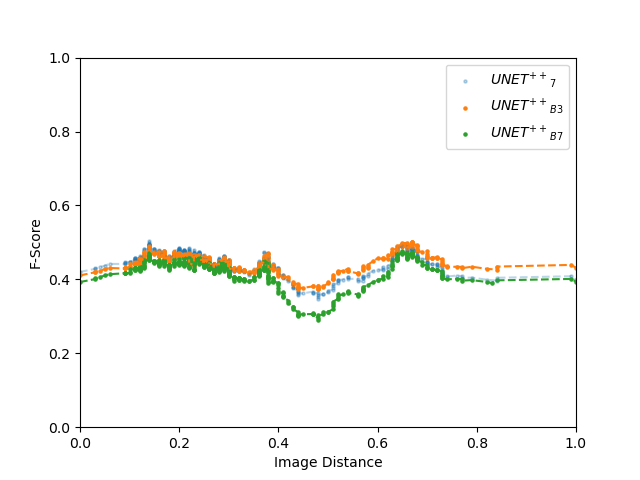}
		\caption{$S_3$}
		\label{fig:nonunetGAPs}
	\end{subfigure}
	\begin{subfigure}[h]{0.22\textwidth}
		\centering
		\includegraphics[width=\linewidth,height=0.15\textheight]{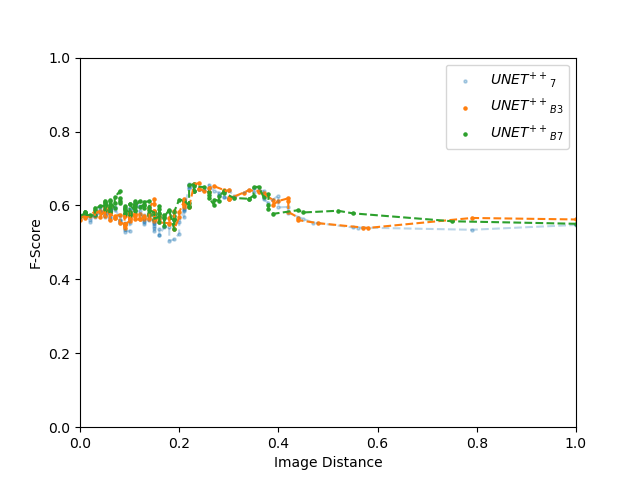}
		\caption{$S_4$}
		\label{fig:nonunetforest}
	\end{subfigure}	
	\caption{This figure shows the  F-score vs $I^{dist}(S, P)$ plot for $S$. The results are computed using the UNet$_7^{++}$, UNet$_{B3}^{++}$ and UNet$_{B7}^{++}$ models.}
	\label{fig:nonunet_if}
\end{figure}

\begin{figure}[htbp!]
	\centering
	\begin{subfigure}[h]{0.22\textwidth}
		\includegraphics[width=\linewidth,height=0.15\textheight]{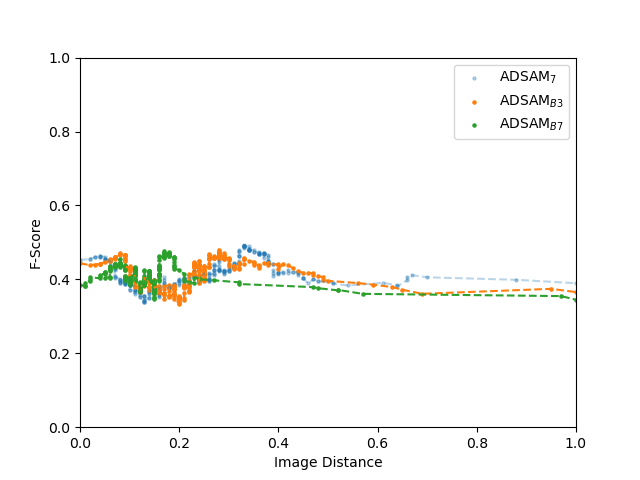}
		\caption{$S_1$}
		\label{fig:nonadsam315}
	\end{subfigure}
	\begin{subfigure}[h]{0.22\textwidth}
		\includegraphics[width=\linewidth,height=0.15\textheight]{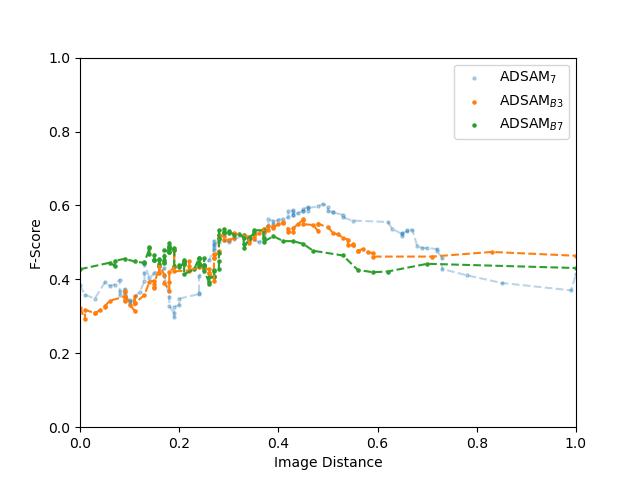}
		\caption{$S_2$}
		\label{fig:nonadsamKWH100}
	\end{subfigure}
	\begin{subfigure}[h]{0.22\textwidth}
		\centering
		\includegraphics[width=\linewidth,height=0.15\textheight]{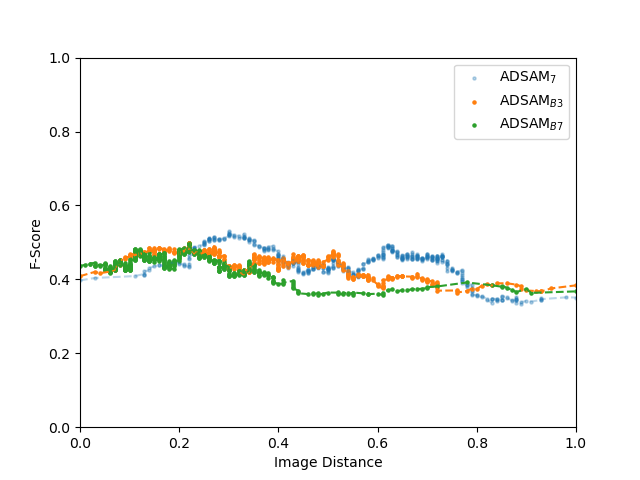}
		\caption{$S_3$}
		\label{fig:nonadsamGAPs}
	\end{subfigure}
	\begin{subfigure}[h]{0.22\textwidth}
		\centering
		\includegraphics[width=\linewidth,height=0.15\textheight]{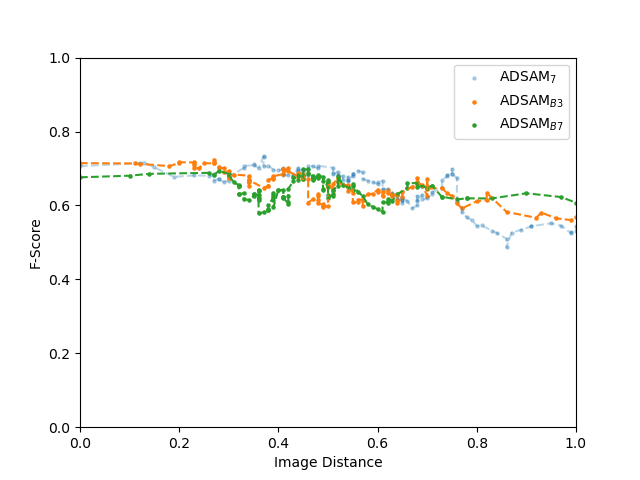}
		\caption{$S_4$}
		\label{fig:nonadsamforest}
	\end{subfigure}	
	\caption{This figure shows the F-score vs $I^{dist}(S, P)$ plot for $S$. The results are computed using the ADSAM$_7$, ADSAM$_{B3}$ and ADSAM$_{B7}$ models.}
	\label{fig:nonadsam_if}
\end{figure}

The results in the figures~\ref{fig:nondc_if},~\ref{fig:nonunet_if} and ~\ref{fig:nonadsam_if} show that the M$_{B3}$ and M$_{B7}$ models (represented by the green and red colored lines) do not suffer from a loss of generalization. The overall F-scores of the  M$_{B3}$ and M$_{B7}$ models are similar to the M$_{7}$ model on the secondary datasets.

\begin{figure}[htbp!]
	\centering
	\includegraphics[width=0.48\textwidth,height=0.15\textheight]{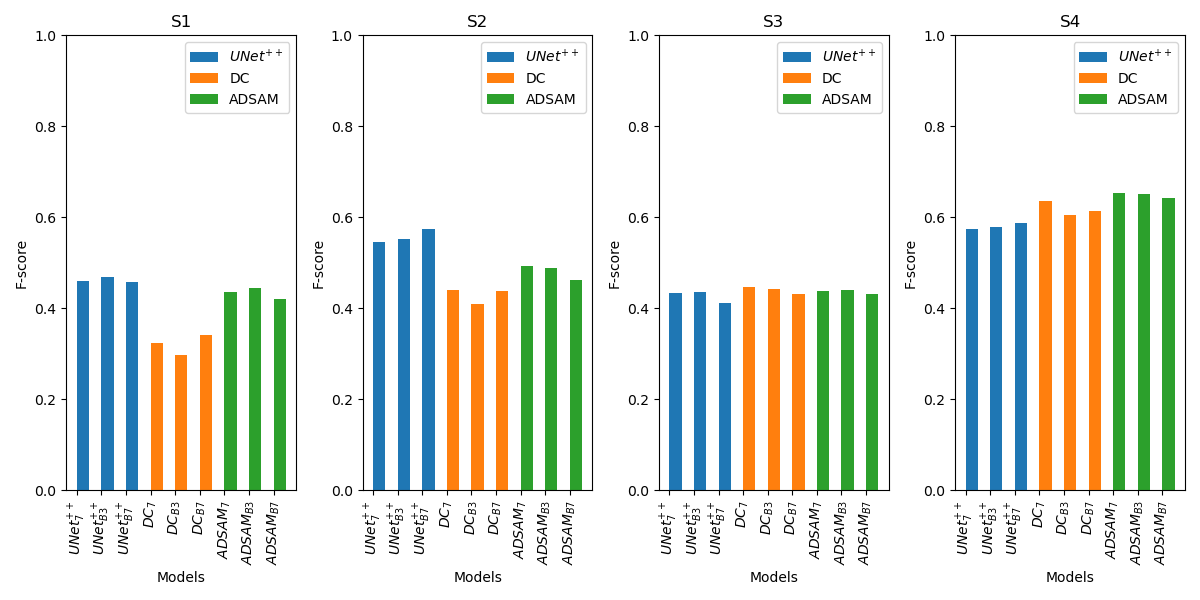}
	\caption{The figure shows the performance (F-score) of the M$_{7}$, M$_{B3}$ and M$_{B7}$ models on the secondary datasets, $S$.}
	\label{fig:nonfdist}
\end{figure}

\section{Scene-Centric Analysis}
\label{scene}
Scene recognition is a challenging task that requires classification of complex and diverse scenes in indoor and outdoor environments. As these images contain diverse objects and backgrounds, it can cause ambiguity in classification~\ref{scenerec}. In this section, we perform experiments to understand the distinguishing capabilities of the models between different scenes using the distance metrics discussed in section~\ref{distancecomp}. We randomly select 50 different scene categories from the partition-1\footnote{\url{https://www.tensorflow.org/datasets/catalog/sun397}} of SUN397~\ref{sunpaper} dataset for the analysis (see table~\ref{tab:singlecell} for details). 

\begin{table}[ht]
	\centering
	\begin{tabular}{|c|}
		\hline
		\parbox{0.9\linewidth}{ % Adjust the width as needed
			\vspace{5pt} 
			\begin{enumerate*}[1.,font=\color{blue},itemjoin={\tab}]
				\item apartment\_building
				\item apse
				\item aquarium
				\item arch
				\item art\_school
				\item badlands
				\item banquet\_hall
				\item basement
				\item beauty\_salon
				\item berth
				\item bookstore
				\item botanical\_garden
				\item butte
				\item cabin
				\item campus
				\item car\_interior
				\item coast
				\item control\_room
				\item covered\_bridge
				\item dentists\_office
				\item driveway
				\item elevator
				\item factory
				\item fastfood\_restaurant
				\item field
				\item fountain
				\item garbage\_dump
				\item hangar
				\item hospital\_room
				\item house
				\item islet
				\item jewelry\_shop
				\item kitchen
				\item kitchenette
				\item marsh
				\item physics\_laboratory
				\item playroom
				\item poolroom
				\item power\_plant
				\item racecourse
				\item restaurant\_kitchen
				\item rice\_paddy
				\item ruin
				\item runway
				\item supermarket
				\item synagogue
				\item toyshop
				\item waiting\_room
				\item wind\_farm
				\item yard
			\end{enumerate*}
		} \\
		\hline
	\end{tabular}
	\caption{SUN397 categories selected for the experiments.}
	\label{tab:singlecell}
\end{table}

The train and test images for each category are resized to $448 \times 448$ and combined into a single set prior to the analysis. This results in 50 scene datasets, each containing 100-200 images. We select images from one of the categories as the primary dataset and compute it's distance with the 49 other categories that are secondary datasets. For this task, apart from ENet, we use 4 other models listed in the table~\ref{table:scenefeat}.

\begin{table}
	\begin{tabular}{lp{5cm}}
		\toprule
		Model & Feature Vector \\
		\midrule		
		CLIPModel & The image features from the CLIP visual encoder~\ref{clipseg} without any text prompt.\\
		DinoV2~\ref{dinov2} & The last hidden state of a base-size vision transformer (encoder model) trained using the DinoV2 method.\\
		ConvNeXtV2~\ref{convnextv2} & The last hidden state of the base-sized model\footnote{\url{https://huggingface.co/facebook/convnextv2-base-22k-224}} finetuned on ImageNet-22k.\\
		SwinV2~\ref{swinv2} & The last hidden state of the base-sized model pre-trained on ImageNet-21k.\\
		\bottomrule
	\end{tabular}
	\caption{The table shows the models used and the details of the corresponding feature vector extracted from it.}
	\label{table:scenefeat}
\end{table}

The closest five and farthest 5 categories in the dataset in terms of O$^{dist}(S, P)$ are obtained to analyze the applicability of the proposed distances on these scene-centric datasets. Figure 18 shows the I$^{dist}$ distribution of the closest and farthest 5 categories from the primary dataset (field category) for the selected models.

\begin{figure*}[pos=htpb!,align=\centering]
	\begin{subfigure}[t]{0.48\textwidth}
		\includegraphics[width=\linewidth,height=0.28\textheight]{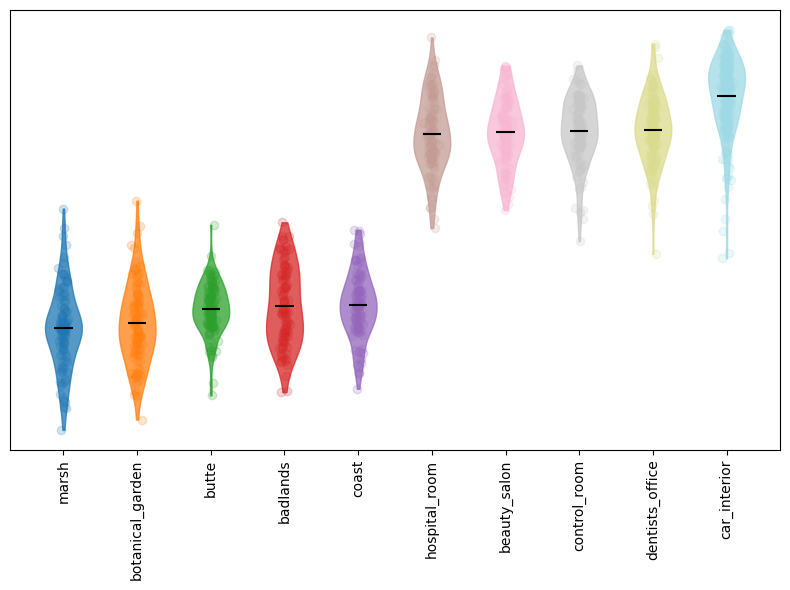}
		\caption{CLIPModel}
		\label{fig:CLIPSegVoilin}
	\end{subfigure}
	\begin{subfigure}[t]{0.48\textwidth}
		\centering
		\includegraphics[width=\linewidth,height=0.28\textheight]{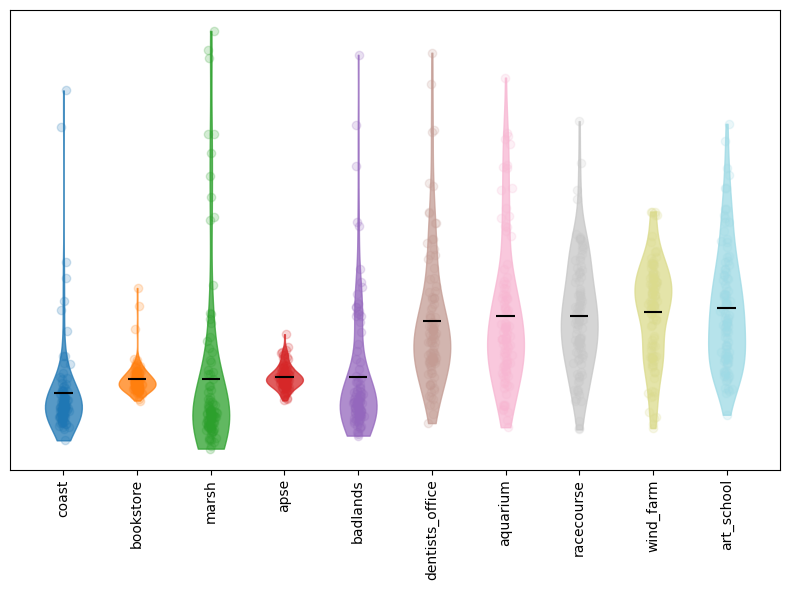}
		\caption{ENet}
		\label{fig:ENetVoilin}
	\end{subfigure}
	\begin{subfigure}[t]{0.48\textwidth}
		\centering
		\includegraphics[width=\linewidth,height=0.28\textheight]{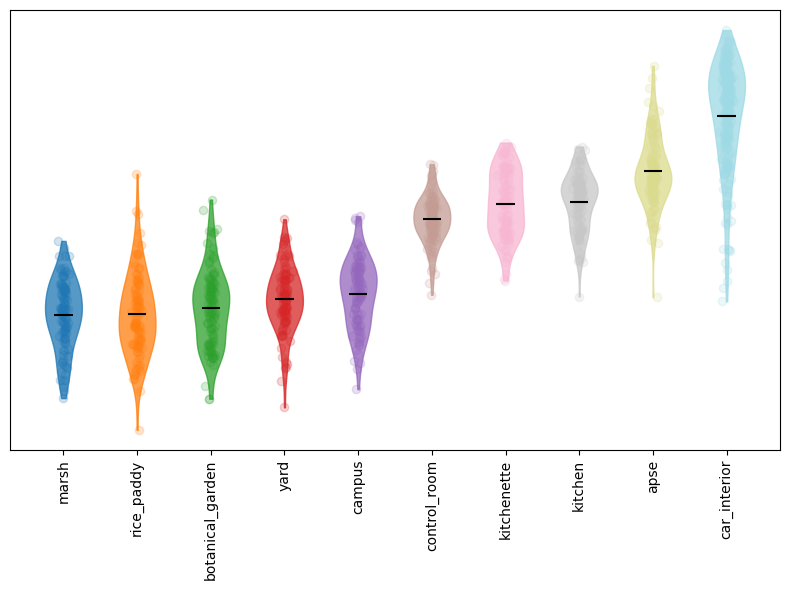}
		\caption{DinoV2}
		\label{fig:DinoV2Voilin}
	\end{subfigure}	
	\begin{subfigure}[t]{0.48\textwidth}
		\centering
		\includegraphics[width=\linewidth,height=0.28\textheight]{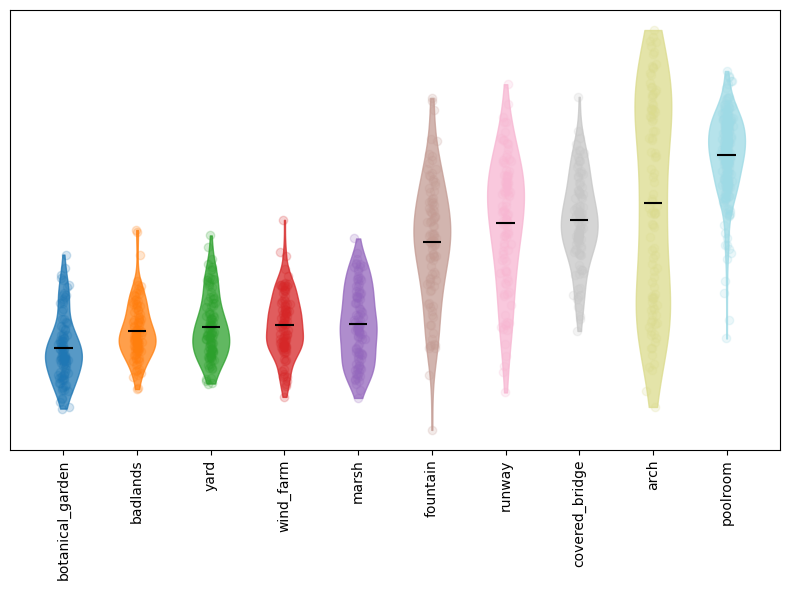}
		\caption{ConvNeXtV2}
		\label{fig:ConvNeXtV2Voilin}
	\end{subfigure}
	\begin{subfigure}[t]{0.48\textwidth}
		\centering
		\includegraphics[width=\linewidth,height=0.28\textheight]{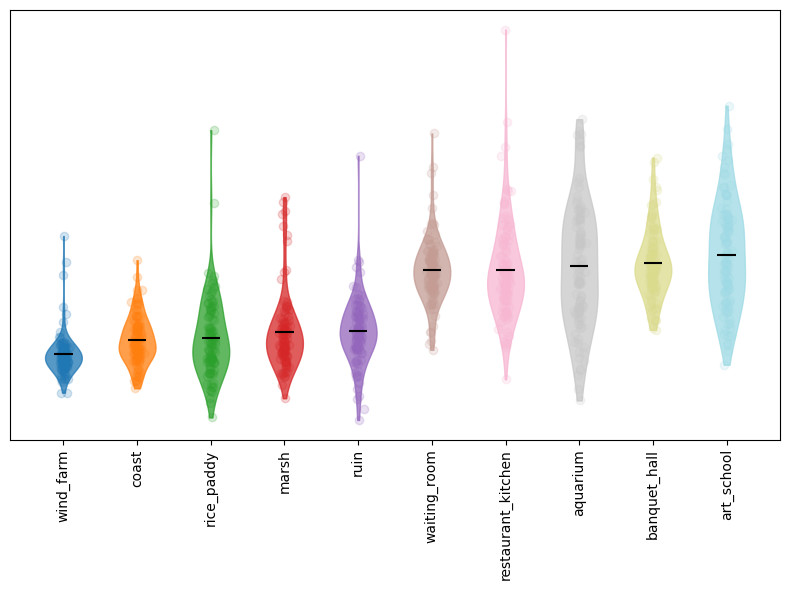}
		\caption{SwinV2}
		\label{fig:SwinV2Voilin}
	\end{subfigure}	
	\caption{This figure shows the voilin plots of I$^{dist}$ from various models. The I$^{dist}$ for each image in the listed categories (see x-axis labels) is computed by taking the \textbf{primary} dataset as \textbf{field}. From left to right, the first 5 voilin plots are for the closest categories and the last 5 voilin plots are for the farthest categories from field.}
	\label{fig:sunvoilin}
\end{figure*}

\begin{figure}[pos=htpb!,align=\centering]
	\begin{subfigure}[t]{0.48\textwidth}
		\includegraphics[width=\linewidth,height=0.15\textheight]{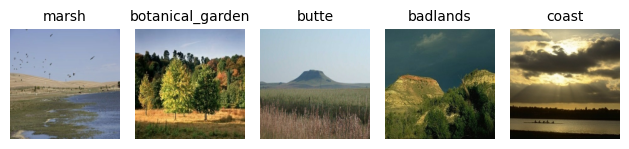}
		\caption{CLIPModel}
		\label{fig:CLIPSegfield}
	\end{subfigure}
	\begin{subfigure}[h]{0.48\textwidth}
		\centering
		\includegraphics[width=\linewidth,height=0.15\textheight]{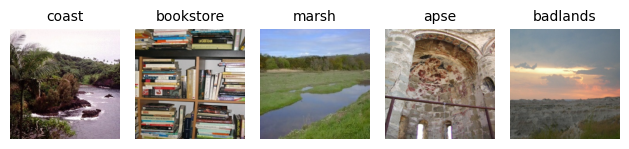}
		\caption{ENet}
		\label{fig:ENetfield}
	\end{subfigure}
	\begin{subfigure}[h]{0.48\textwidth}
		\centering
		\includegraphics[width=\linewidth,height=0.15\textheight]{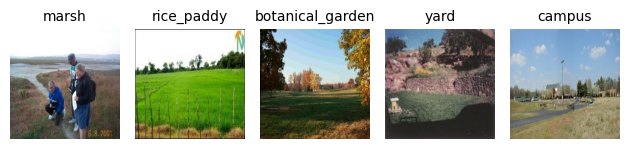}
		\caption{DinoV2}
		\label{fig:DinoV2field}
	\end{subfigure}	
	\begin{subfigure}[h]{0.48\textwidth}
		\centering
		\includegraphics[width=\linewidth,height=0.15\textheight]{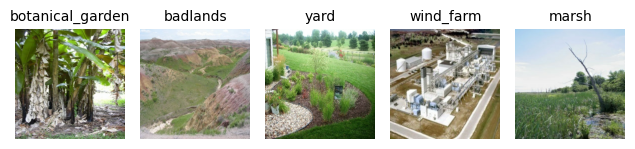}
		\caption{ConvNeXtV2}
		\label{fig:ConvNeXtV2field}
	\end{subfigure}
	\begin{subfigure}[h]{0.48\textwidth}
		\centering
		\includegraphics[width=\linewidth,height=0.15\textheight]{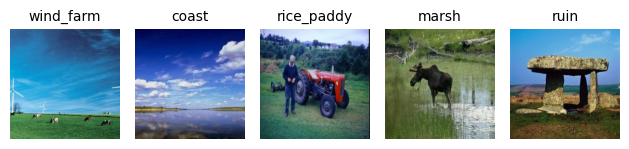}
		\caption{SwinV2}
		\label{fig:SwinV2field}
	\end{subfigure}	
	\caption{This figure shows the closest images, based on I$^{dist}(S, P)$, where P is \textbf{field} for the selected models.}
	\label{fig:sunfield}
\end{figure}

\begin{figure}[pos=htpb!,align=\centering]
	\begin{subfigure}[t]{0.48\textwidth}
		\includegraphics[width=\linewidth,height=0.15\textheight]{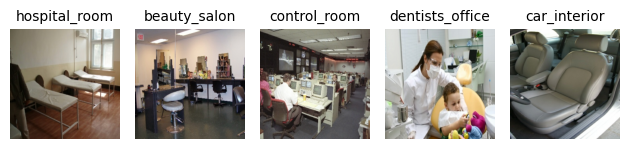}
		\caption{CLIPModel}
		\label{fig:CLIPSegfieldfarthest}
	\end{subfigure}
	\begin{subfigure}[h]{0.48\textwidth}
		\centering
		\includegraphics[width=\linewidth,height=0.15\textheight]{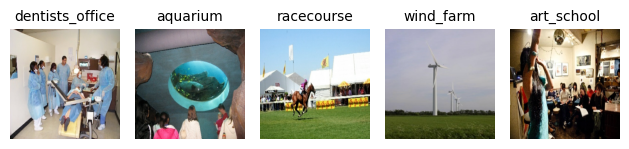}
		\caption{ENet}
		\label{fig:ENetfieldfarthest}
	\end{subfigure}
	\begin{subfigure}[h]{0.48\textwidth}
		\centering
		\includegraphics[width=\linewidth,height=0.15\textheight]{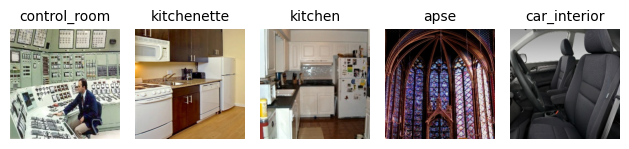}
		\caption{DinoV2}
		\label{fig:DinoV2fieldfarthest}
	\end{subfigure}	
	\begin{subfigure}[h]{0.48\textwidth}
		\centering
		\includegraphics[width=\linewidth,height=0.15\textheight]{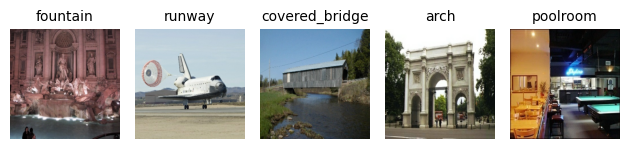}
		\caption{ConvNeXtV2}
		\label{fig:ConvNeXtV2fieldfarthest}
	\end{subfigure}
	\begin{subfigure}[h]{0.48\textwidth}
		\centering
		\includegraphics[width=\linewidth,height=0.15\textheight]{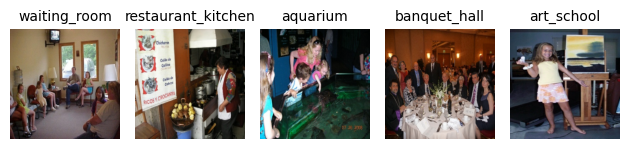}
		\caption{SwinV2}
		\label{fig:SwinV2fieldfarthest}
	\end{subfigure}	
	\caption{This figure shows the farthest images, based on I$^{dist}(S, P)$, where P is \textbf{field} for the selected models.}
	\label{fig:sunfieldfarthest}
\end{figure}

\begin{figure}[pos=htpb!,align=\centering]
	\begin{subfigure}[t]{0.48\textwidth}
		\includegraphics[width=\linewidth,height=0.15\textheight]{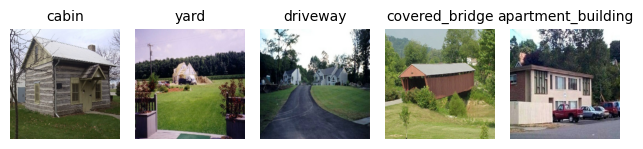}
		\caption{CLIPModel}
		\label{fig:CLIPSeghouse}
	\end{subfigure}
	\begin{subfigure}[h]{0.48\textwidth}
		\centering
		\includegraphics[width=\linewidth,height=0.15\textheight]{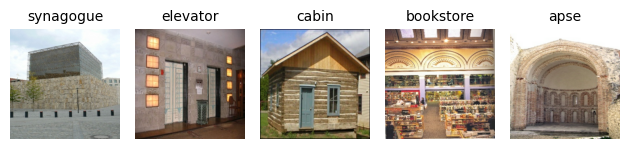}
		\caption{ENet}
		\label{fig:ENethouse}
	\end{subfigure}
	\begin{subfigure}[h]{0.48\textwidth}
		\centering
		\includegraphics[width=\linewidth,height=0.15\textheight]{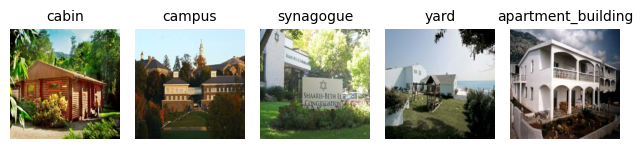}
		\caption{DinoV2}
		\label{fig:DinoV2house}
	\end{subfigure}	
	\begin{subfigure}[h]{0.48\textwidth}
		\centering
		\includegraphics[width=\linewidth,height=0.15\textheight]{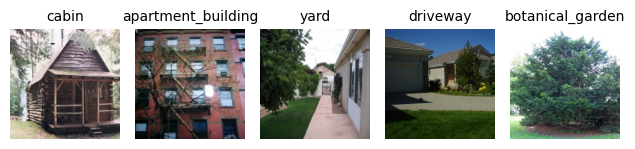}
		\caption{ConvNeXtV2}
		\label{fig:ConvNeXtV2house}
	\end{subfigure}
	\begin{subfigure}[h]{0.48\textwidth}
		\centering
		\includegraphics[width=\linewidth,height=0.15\textheight]{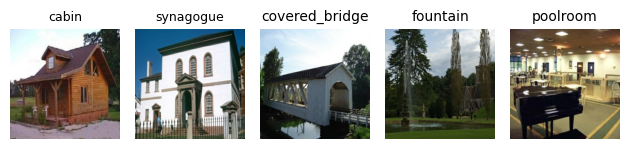}
		\caption{SwinV2}
		\label{fig:SwinV2house}
	\end{subfigure}	
	\caption{This figure shows the closest images, based on I$^{dist}(S, P)$, where P is \textbf{house} for the selected models.}
	\label{fig:sunhouse}
\end{figure}

\begin{figure}[pos=htpb!,align=\centering]
	\begin{subfigure}[t]{0.48\textwidth}
		\includegraphics[width=\linewidth,height=0.15\textheight]{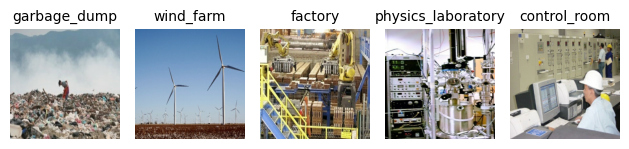}
		\caption{CLIPModel}
		\label{fig:CLIPSeghousefarthest}
	\end{subfigure}
	\begin{subfigure}[h]{0.48\textwidth}
		\centering
		\includegraphics[width=\linewidth,height=0.15\textheight]{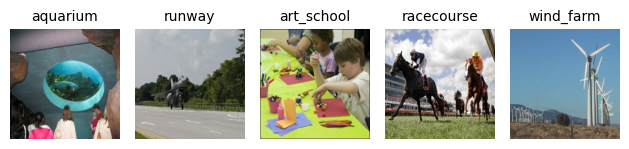}
		\caption{ENet}
		\label{fig:ENethousefarthest}
	\end{subfigure}
	\begin{subfigure}[h]{0.48\textwidth}
		\centering
		\includegraphics[width=\linewidth,height=0.15\textheight]{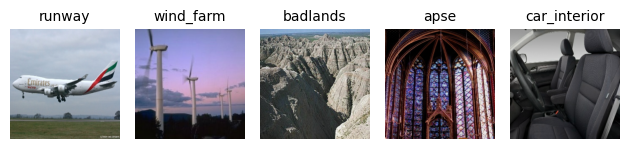}
		\caption{DinoV2}
		\label{fig:DinoV2housefarthest}
	\end{subfigure}	
	\begin{subfigure}[h]{0.48\textwidth}
		\centering
		\includegraphics[width=\linewidth,height=0.15\textheight]{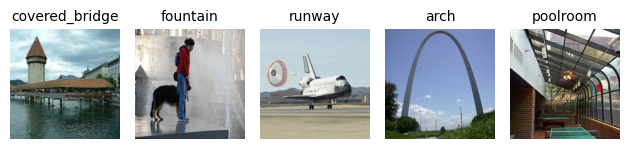}
		\caption{ConvNeXtV2}
		\label{fig:ConvNeXtV2housefarthest}
	\end{subfigure}
	\begin{subfigure}[h]{0.48\textwidth}
		\centering
		\includegraphics[width=\linewidth,height=0.15\textheight]{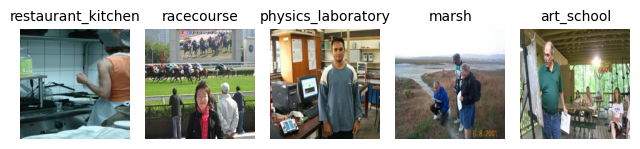}
		\caption{SwinV2}
		\label{fig:SwinV2housefarthest}
	\end{subfigure}	
	\caption{This figure shows the farthest images, based on I$^{dist}(S, P)$, where P is \textbf{house} for the selected models.}
	\label{fig:sunhousefarthest}
\end{figure}

\begin{figure}[pos=htpb!,align=\centering]
	\begin{subfigure}[t]{0.45\textwidth}
		\includegraphics[width=\linewidth,height=0.11\textheight]{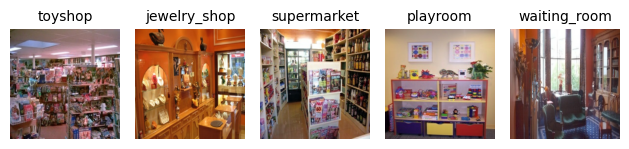}
		\caption{CLIPModel}
		\label{fig:CLIPSegbookstoreclosest}
	\end{subfigure}
	\begin{subfigure}[h]{0.45\textwidth}
		\centering
		\includegraphics[width=\linewidth,height=0.11\textheight]{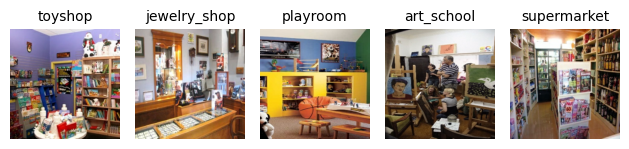}
		\caption{DinoV2}
		\label{fig:DinoV2bookstoreclosest}
	\end{subfigure}
	\caption{This figure shows the closest images, based on I$^{dist}(S, P)$, where P is \textbf{bookstore} for the selected models.}
	\label{fig:sunbookstoreclosest}
\end{figure}

\begin{figure}[pos=htpb!,align=\centering]
	\begin{subfigure}[t]{0.45\textwidth}
		\includegraphics[width=\linewidth,height=0.11\textheight]{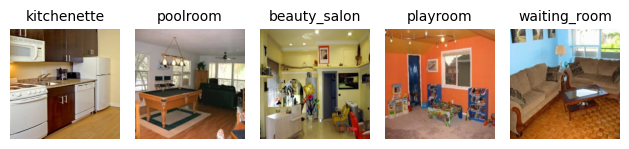}
		\caption{CLIPModel}
		\label{fig:CLIPkitchenclosest}
	\end{subfigure}
	\begin{subfigure}[h]{0.45\textwidth}
		\centering
		\includegraphics[width=\linewidth,height=0.11\textheight]{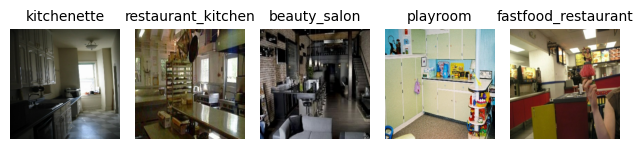}
		\caption{DinoV2}
		\label{fig:DinoV2kitchenclosest}
	\end{subfigure}
	\caption{This figure shows the closest images, based on I$^{dist}(S, P)$, where P is \textbf{kitchen} for the selected models.}
	\label{fig:sunkitchenclosest}
\end{figure}

\begin{figure}[pos=htpb!,align=\centering]
	\begin{subfigure}[t]{0.45\textwidth}
		\includegraphics[width=\linewidth,height=0.11\textheight]{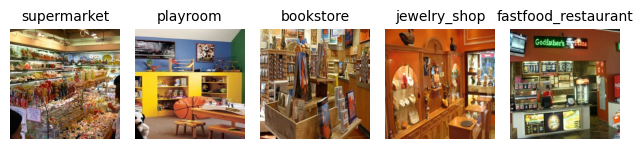}
		\caption{CLIPModel}
		\label{fig:CLIPtoyshopclosest}
	\end{subfigure}
	\begin{subfigure}[h]{0.45\textwidth}
		\centering
		\includegraphics[width=\linewidth,height=0.11\textheight]{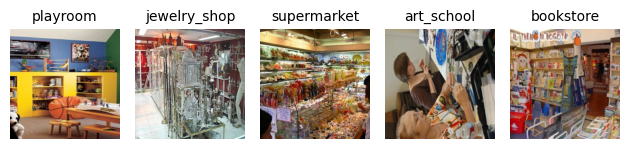}
		\caption{DinoV2}
		\label{fig:DinoV2toyshopclosest}
	\end{subfigure}
	\caption{This figure shows the closest images, based on I$^{dist}(S, P)$, where P is \textbf{toyshop} for the selected models.}
	\label{fig:suntoyshopclosest}
\end{figure}

\begin{figure}[pos=htpb!,align=\centering]
	\begin{subfigure}[t]{0.45\textwidth}
		\includegraphics[width=\linewidth,height=0.11\textheight]{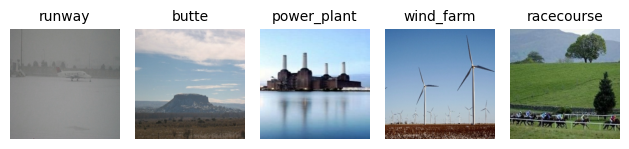}
		\caption{CLIPModel}
		\label{fig:CLIPbookstorefarthest}
	\end{subfigure}
	\begin{subfigure}[h]{0.45\textwidth}
		\centering
		\includegraphics[width=\linewidth,height=0.11\textheight]{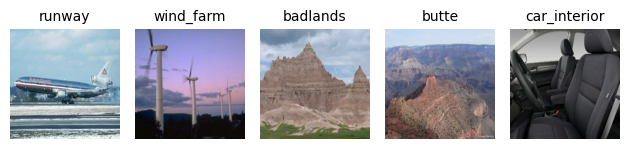}
		\caption{DinoV2}
		\label{fig:DinoV2bookstorefarthest}
	\end{subfigure}
	\caption{This figure shows the farthest images, based on I$^{dist}(S, P)$, where P is \textbf{bookstore} for the selected models.}
	\label{fig:sunbookstorefarthest}
\end{figure}

\begin{figure}[pos=htpb!,align=\centering]
		\begin{subfigure}[t]{0.45\textwidth}
			\includegraphics[width=\linewidth,height=0.11\textheight]{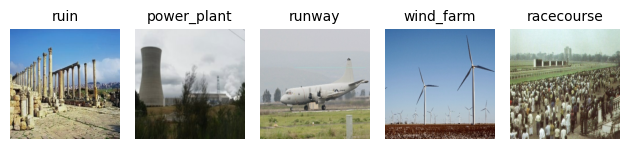}
			\caption{CLIPModel}
			\label{fig:CLIPkitchenfarthest}
		\end{subfigure}
		\begin{subfigure}[h]{0.45\textwidth}
			\centering
			\includegraphics[width=\linewidth,height=0.11\textheight]{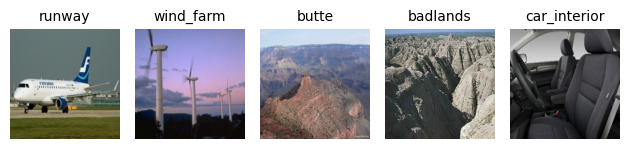}
			\caption{DinoV2}
			\label{fig:DinoV2kitchenfarthest}
		\end{subfigure}
		\caption{This figure shows the farthest images, based on I$^{dist}(S, P)$, where P is \textbf{kitchen} for the selected models.}
		\label{fig:sunkitchenfarthest}
	\end{figure}

\begin{figure}[pos=htpb!,align=\centering]
		\begin{subfigure}[t]{0.45\textwidth}
			\includegraphics[width=\linewidth,height=0.11\textheight]{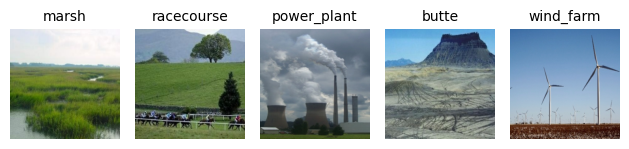}
			\caption{CLIPModel}
			\label{fig:CLIPSegtoyshopfarthest}
		\end{subfigure}
		\begin{subfigure}[h]{0.45\textwidth}
			\centering
			\includegraphics[width=\linewidth,height=0.11\textheight]{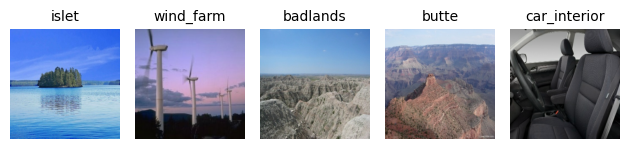}
			\caption{DinoV2}
			\label{fig:DinoV2toyshopfarthest}
		\end{subfigure}
		\caption{This figure shows the farthest images, based on I$^{dist}(S, P)$, where P is \textbf{toyshop} for the selected models.}
		\label{fig:suntoyshopfarthest}
	\end{figure}

In the qualitative analysis, we obtain the closest and farthest 5 images (1 taken from each secondary dataset) from the primary dataset based on the I$^{dist}$. The difference in semantics of the closest and farthest images helps in understanding the image features extracted by the models. Figures~\ref{fig:sunfield},~\ref{fig:sunfieldfarthest},~\ref{fig:sunhouse} and~\ref{fig:sunhousefarthest} show that if the primary dataset is outdoor, the closest datasets are closely related outdoor scenes. We investigate the distances computed by these models for indoor primary datasets and obtain the closest and farthest 5 categories. We observe that CLIPModel and DinoV2 provide meaningful closest and farthest scenes. This can be seen in figures~\ref{fig:sunbookstoreclosest},~\ref{fig:sunkitchenclosest},~\ref{fig:suntoyshopclosest},~\ref{fig:sunbookstorefarthest},~\ref{fig:sunkitchenfarthest} and~\ref{fig:suntoyshopfarthest}. The primary categories of bookstore, kitchen and toyshop are matched with closely related indoor scenes and are far from the outdoor categories.

Taking the field category as the primary dataset, we repeat the experiments by adding Gaussian Noise and changing brightness to examine the robustness of features of the models. Firstly, we increase the brightness of all the images of the closest and farthest 5 categories by 30\% and compute the distances of all secondary datasets from the primary dataset (field category). Secondly, we decrease the brightness of these images by 30\% and repeat the same process. Finally, we apply Gaussian blur for adding noise to these images and compute the distances. We observe that there are slight changes in the order of the closest 5 categories for the CLIPModel, DinoV2, and SwinV2 models whereas the ENet and ConvNeXtV2 models are impacted the most by the change in brightness and blur as the closest and farthest categories are changed. Overall, features like brightness and noise do affect the feature representations of the models.

\begin{figure*}[pos=htpb!,align=\centering]
	\begin{subfigure}[t]{0.95\textwidth}
		\includegraphics[width=\linewidth,height=0.13\textheight]{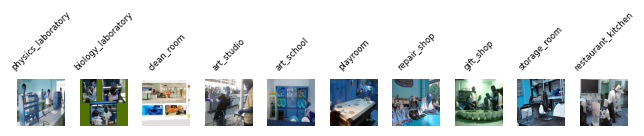}
		\caption{DinoV2, P is chemistry\_lab; The closest categories are physics\_laboratory, biology\_laboratory and dean\_room scenes.}
		\label{fig:clDino}
	\end{subfigure}
	\begin{subfigure}[t]{0.95\textwidth}
		\includegraphics[width=\linewidth,height=0.13\textheight]{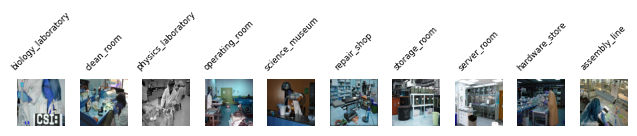}
		\caption{CLIPModel, P is chemistry\_lab; The closest categories are physics\_laboratory, dean\_room, biology\_laboratory scenes.}
		\label{fig:clCLIP}
	\end{subfigure}
	\begin{subfigure}[h]{0.95\textwidth}
		\centering
		\includegraphics[width=\linewidth,height=0.13\textheight]{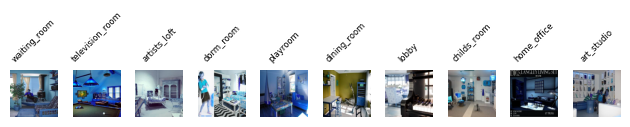}
		\caption{DinoV2, P is living\_room; The closest categories are waiting\_room, television\_room and artists\_loft scenes.}
		\label{fig:lrDino}
	\end{subfigure}
	\begin{subfigure}[h]{0.95\textwidth}
		\centering
		\includegraphics[width=\linewidth,height=0.13\textheight]{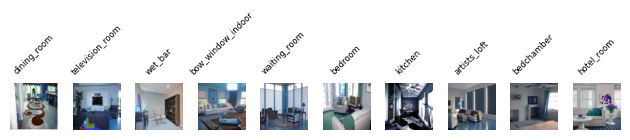}
		\caption{CLIPModel, P is living\_room; The closest categories are dining\_room, television\_room and wet\_bar scenes.}
		\label{fig:lrCLIP}
	\end{subfigure}	
	\begin{subfigure}[h]{0.95\textwidth}
		\centering
		\includegraphics[width=\linewidth,height=0.13\textheight]{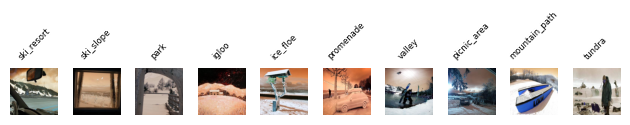}
		\caption{DinoV2, P is snowfield; The closest categories are ski\_resort, ski\_slope and park scenes.}
		\label{fig:snDino}
	\end{subfigure}
	\begin{subfigure}[h]{0.95\textwidth}
		\centering
		\includegraphics[width=\linewidth,height=0.13\textheight]{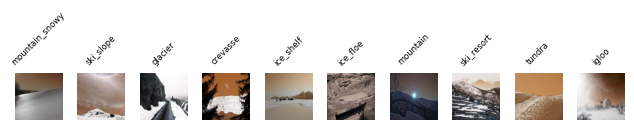}
		\caption{CLIPModel, P is snowfield; The closest categories are mountain\_snowy, ski\_slope and glacier scenes.}
		\label{fig:snCLIP}
	\end{subfigure}	
	\caption{This figure shows the closest images, based on I$^{dist}(S, P)$, where P is mentioned in the caption for the CLIP and DinoV2 models.}
	\label{fig:placeclidino}
\end{figure*}

Based on these experiments, we find the CLIPModel and DinoV2 to be more discriminative in both indoor and outdoor scenes as compared to other models. Therefore, to further examine the extent of discriminative nature of these models, we select the Places365~\ref{places} training set\footnote{\url{https://www.kaggle.com/datasets/nickj26/places2-mit-dataset}} with about 5000 images for all categories and consider indoor and outdoor scene categories as primary datasets. Figure~\ref{fig:placeclidino} shows the closest 10 categories for the CLIPModel and DinoV2 models using images from chemistry\_lab, living\_room and snowfield respectively, as the primary datasets . Overall, it can be seen that the closest categories are meaningful i.e semantically similar scenes are closer to the corresponding primary scenes.

\section{Person Re-Identification and Distances}
\label{personreid}

\begin{figure}[pos=htpb!,align=\centering]
	\begin{subfigure}[t]{0.2\textwidth}
		\includegraphics[width=\linewidth,height=0.12\textheight]{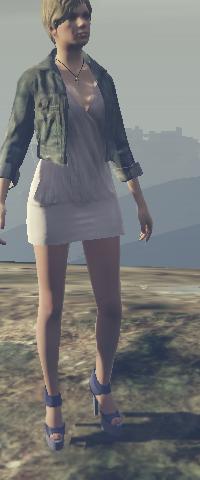}
		\caption{Person 1}
		\label{fig:p1}
	\end{subfigure}
	\begin{subfigure}[t]{0.2\textwidth}
		\centering
		\includegraphics[width=\linewidth,height=0.12\textheight]{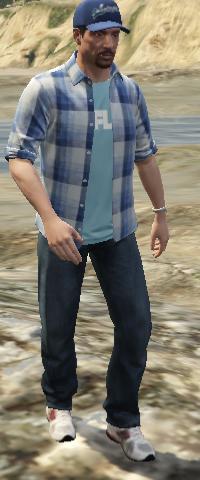}
		\caption{Person 2}
		\label{fig:p3}
	\end{subfigure}
	\caption{This figure shows the images of two persons from the FineGPR person re-identification subset.}
	\label{fig:person}
\end{figure}

\begin{figure}[pos=htpb!,align=\centering]
 	\begin{subfigure}[t]{0.45\textwidth}
 		\includegraphics[width=\linewidth,height=0.11\textheight]{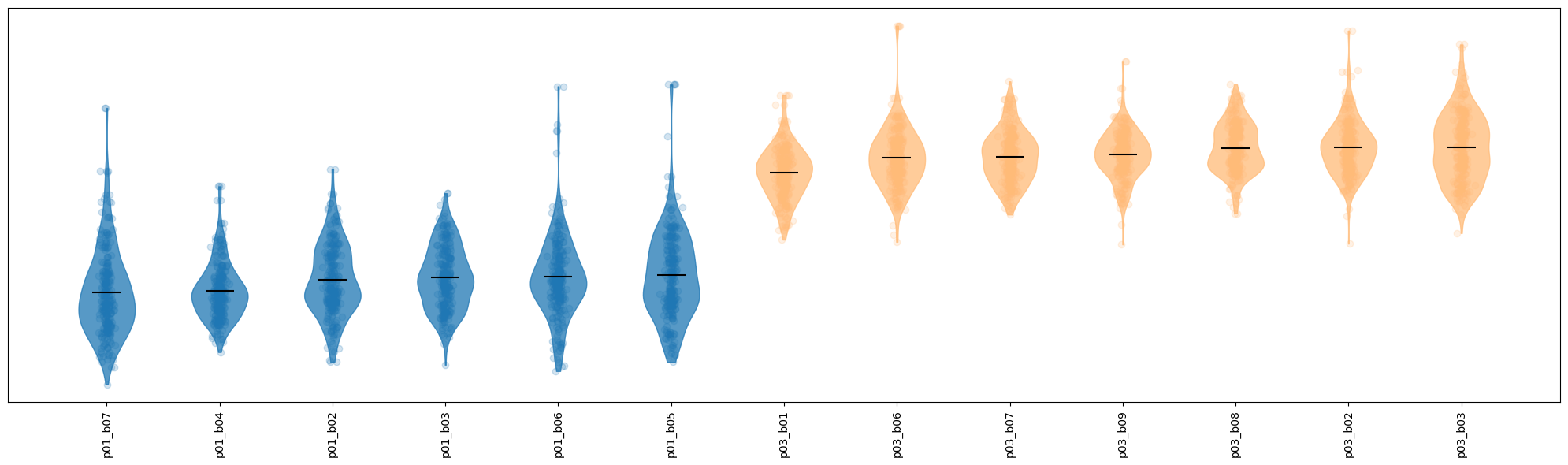}
 		\caption{The I$^{dist}$ voilin plots of all person-scene combination sets by taking the primary set as $p_1s_1$. The distances are obtained from ConvNeXtV2 model.}
 		\label{fig:bkgconvnextv2}
 	\end{subfigure}
 	\begin{subfigure}[h]{0.45\textwidth}
 		\centering
 		\includegraphics[width=\linewidth,height=0.11\textheight]{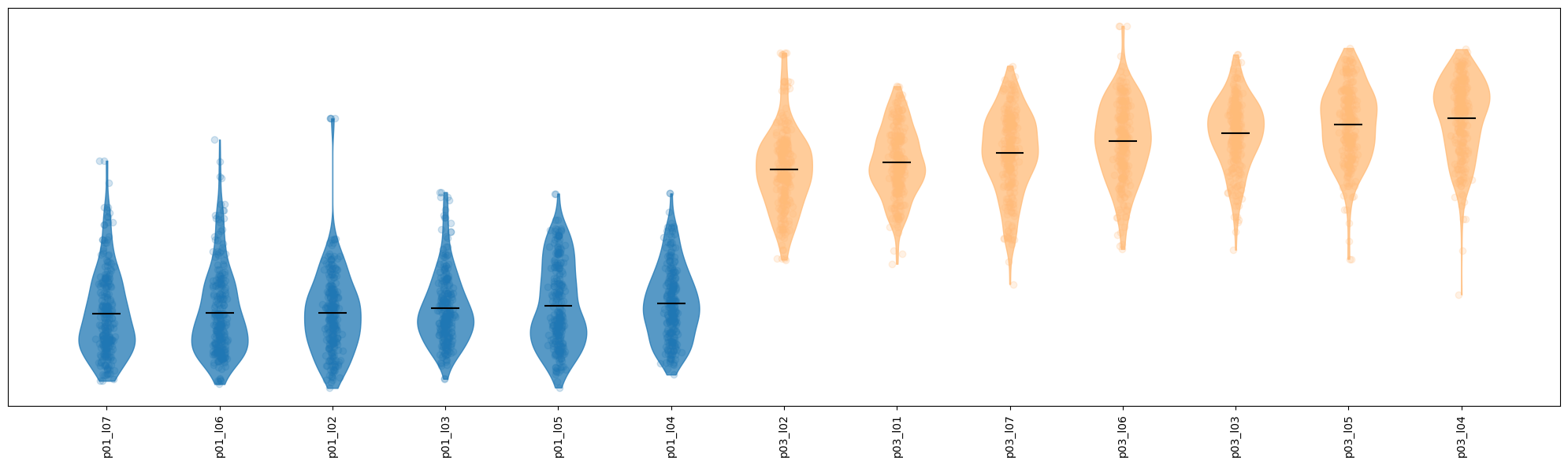}
 		\caption{The I$^{dist}$ voilin plots of all person-weather combination sets by taking the primary set as $p_1s_1$. The distances are obtained from SwinV2 model.}
 		\label{fig:illuminationswinv2}
 	\end{subfigure}
 	\begin{subfigure}[h]{0.45\textwidth}
 		\centering
 		\includegraphics[width=\linewidth,height=0.11\textheight]{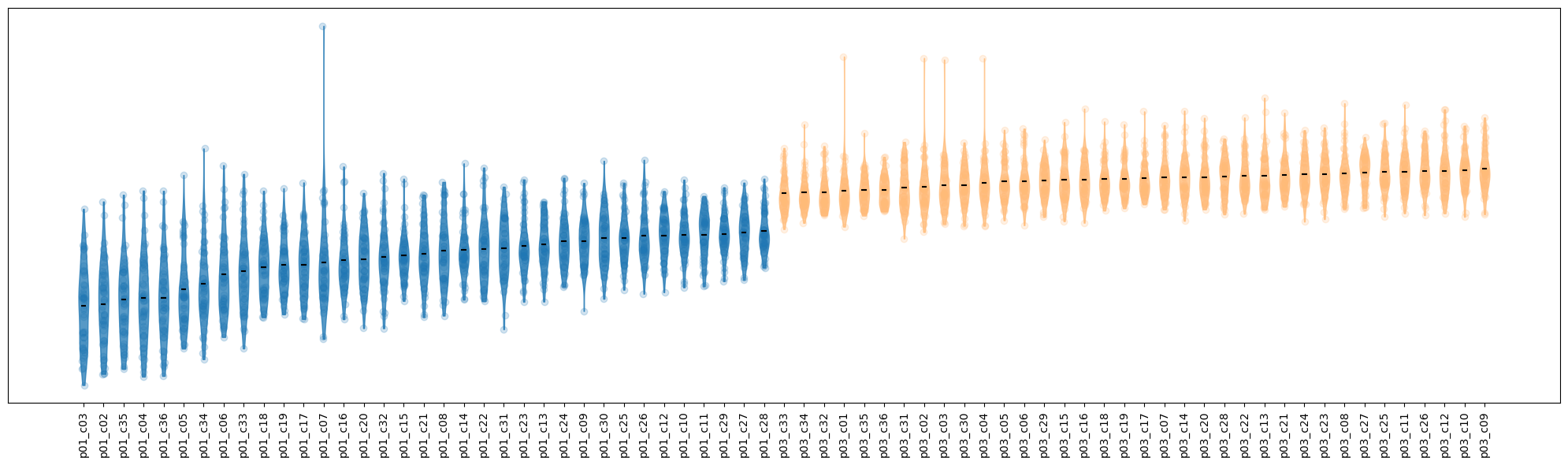}
 		\caption{The I$^{dist}$ voilin plots of all person-viewpoint combination sets by taking the primary set as $p_1w_1$. The distances are obtained from DinoV2 model.}
 		\label{fig:viewpointdinov2}
 	\end{subfigure}
 	\caption{This figure shows the voilin plots of two persons (1,2) captured under different (i) scenes , (ii) illuminations and (iii) viewpoints. The blue colored plots correspond to person 1 and the orange colored plots correspond to person 2.}
 	\label{fig:varvoilin}
 \end{figure}

\begin{figure}[pos=H,align=\centering]
 	\begin{subfigure}[t]{0.45\textwidth}
 		\includegraphics[width=\linewidth,height=0.11\textheight]{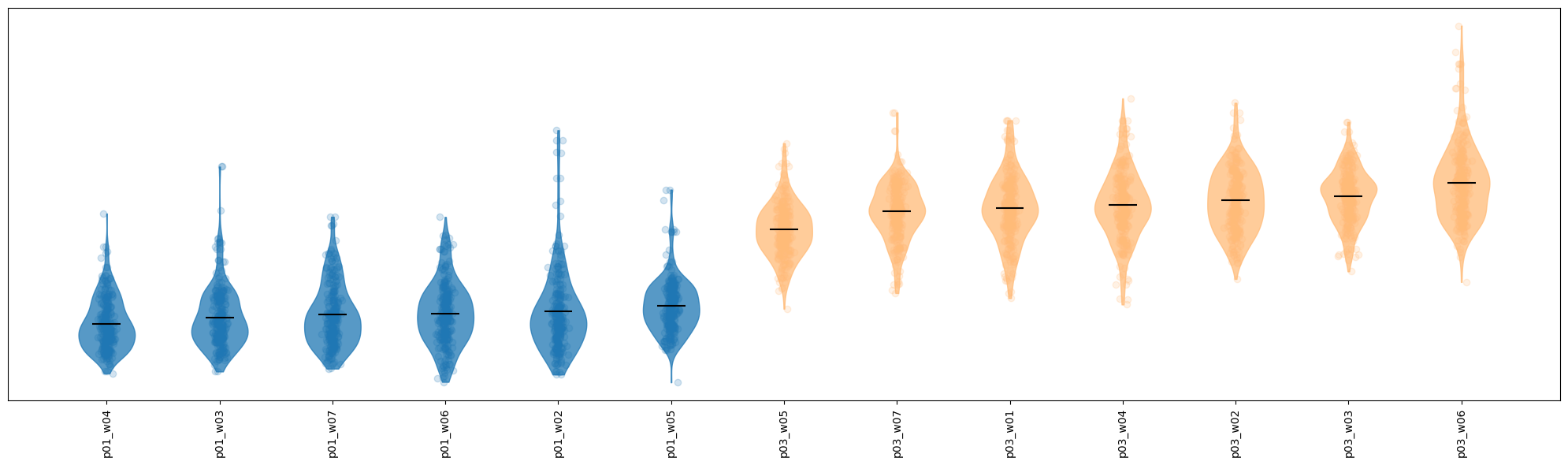}
 		\caption{Without Gaussian Blur}
 		\label{fig:woweatherenet}
 	\end{subfigure}
 	\begin{subfigure}[h]{0.45\textwidth}
 		\centering
 		\includegraphics[width=\linewidth,height=0.11\textheight]{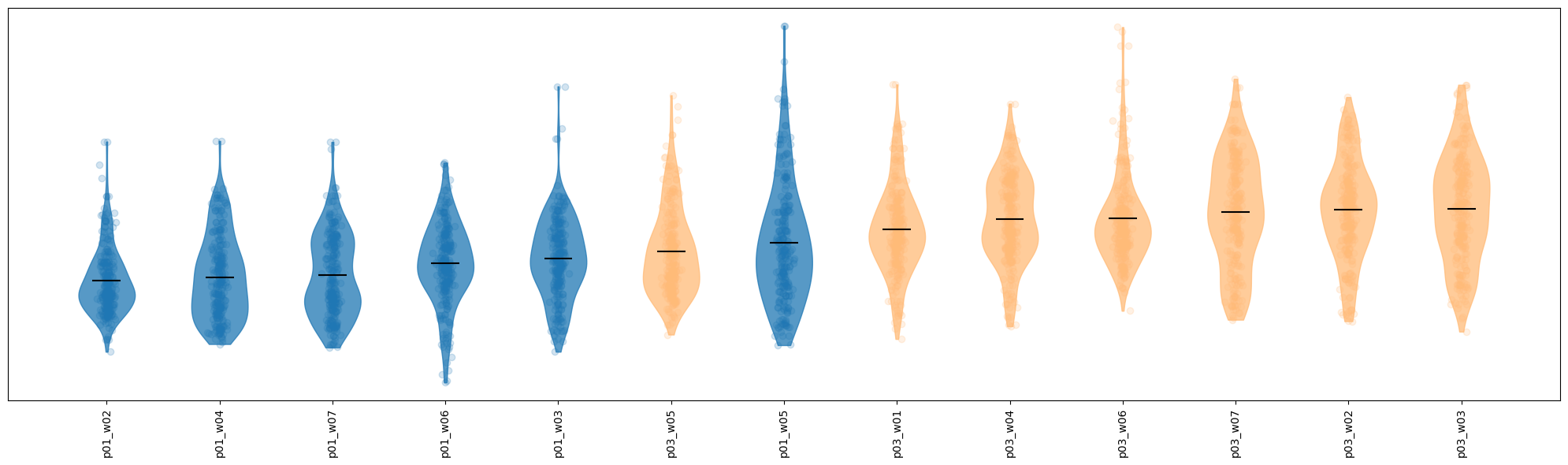}
 		\caption{With Gaussian Blur}
 		\label{fig:wwweatherenet}
 	\end{subfigure}
 	\caption{This figure shows the voilin plots of two persons (1,2) captured under different weather conditions. The blue colored plots correspond to person 1 and the orange colored plots correspond to person 2. The results correspond to the ENet model before and after adding the Gaussian blur. It can be see that the model has become less discriminate from (i) to (ii).}
 	\label{fig:weatherenetvoilin}
 \end{figure}

Person Re-Identification is a task that requires a system to identify a person across different scenes and other variations. The task is solved using many different approaches namely, deep metric learning, local feature learning, generative adversarial learning and sequence feature learning (detailed discussions in~\ref{reidsurvey}). In real-world scenarios, there are significantly many variations and it is important to have a significant diversity in the training data. There are several real-world~\ref{market},~\ref{cuhk03} and synthetic~\ref{unreal},~\ref{synperson} person re-identification datasets. However, we use this\footnote{\url{https://github.com/JeremyXSC/FineGPR/}} version of the FineGPR~\ref{reid} dataset due to its rich diversity and suitability for these experiments. It consists of 1764 images of two persons are provided with different variations mentioned in the table~\ref{table:variabilityfinegpr}. Example images of the two persons are shown in the figure~\ref{fig:person}.
We examine if the models can distinguish between images of two different persons captured under these variations. There are 252 images for each scene, illumination and weather variation and, 49 images for each camera angle/viewpoint variation. In this experimental setup, we fix one kind of variation for both persons and compute the distances O$^{dist}(p_1v_2, p_1v_1)$, O$^{dist}(p_1v_3, p_1v_1)$, ...,  O$^{dist}(p_2v_1, p_1v_1)$,  O$^{dist}(p_2v_2, p_1v_1)$ and so on where p refers to person and v refers to variation. Here, $p_1v_1$ refers to the set of images of person 1 with a fixed variation $v_1$. Say, the set of images under sunny weather of person 1 is fixed while other variations (scenes, illumination and viewpoints) are changing. On computing these distances, we observe that the selected models can distinguish between the two persons. Following are our observations:
\begin{enumerate}
	\item \textbf{Viewpoints:} The dataset has 36 different viewpoints each for person 1 and person 2 listed in the table \ref{table:variabilityfinegpr}. We observe that when we fix person and a specific viewpoint (say, person-1 and viewpoint-1) we found that all other viewpoints of person-1 are closer as compared to all viewpoints of person-2. This indicates that the models can discriminate persons in varying viewpoints. Figure~\ref{fig:viewpointdinov2} shows the corresponding results from the DinoV2 model.	\begin{align}
		\begin{split}
	O^{dist}(p_1a_k, p_1a_1) < O^{dist}(p_2a_j, p_1a_1) \\
	 \forall k \in \{2,3,...,35,36\}, j \in \{1,2,3,..., 35,36\}
	 \end{split}
	 \label{viewpoints}
	\end{align}
	\item \textbf{Scenes:}
	The dataset has 7 different scenes each for person 1 and person 2 listed in the table \ref{table:variabilityfinegpr}. We observe that when we fix person and a specific scene (say, person-1 and scene-1) we found that all other scenes of person-1 are closer as compared to all scenes of person-2. This indicates that the models can discriminate persons in varying scenes. Figure~\ref{fig:bkgconvnextv2} shows the corresponding results from the ConvNeXtV2 model.
	 \begin{align}
		\begin{split}
			O^{dist}(p_1s_k, p_1s_1) < O^{dist}(p_2s_j, p_1s_1) \\
			\forall k \in \{2,3,4,5,6,7\}, j \in \{1,2,3,6,7,8,9\}
		\end{split}
		\label{backgrounds}
	\end{align}
	\item \textbf{Weather \& Illumination:} We observe that when we fix a person and a specific weather/illumination (say, person-1 and weather/illumination-1) we find that all other weather/illumination conditions of person-1 are closer as compared to all weather/illumination conditions of person-2. This indicates that the models can discriminate persons in varying weather/illumination conditions. Figure~\ref{fig:weatherenetvoilin} shows the corresponding results from the SwinV2 model. \begin{align}
		\begin{split}
			O^{dist}(p_1w_k, p_1w_1) < O^{dist}(p_2w_j, p_1w_1) \\
			\forall k \in \{2,3,4,5,6,7\}, j \in \{1,2,3,4,5,6,7\}
		\end{split}
		\label{weaill}
	\end{align}
\end{enumerate}

\begin{table}[H]
	\begin{tabular}{p{3cm}p{3.8cm}}
		\toprule
		Variations&Details\\
		\midrule
		Viewpoints (angles in degrees)&$\{0, 10,..., 340, 350\}$\\
		\midrule
		Scenes& A total of 9 different Urban and Wild scenes\\
		\midrule
		Weather&\parbox{0.9\linewidth}{ % Adjust the width as needed
			\vspace{0.1cm}
			\begin{enumerate*}[1.,font=\color{blue},itemjoin={\tab}]
				\item Sunny
				\item Clouds
				\item Overcast
				\item Foggy
				\item Neutral
				\item Blizzard
				\item Snowlight
			\end{enumerate*}
		} \\
		\midrule
		Illumination&\parbox{0.9\linewidth}{ % Adjust the width as needed
			\vspace{0.2cm}
			\begin{enumerate*}[1.,font=\color{blue},itemjoin={\tab}]
				\item Midnight
				\item Dawn
				\item Forenoon
				\item Noon
				\item Afternoon
				\item Dusk
				\item Night
			\end{enumerate*}
		} \\
		\bottomrule
	\end{tabular}
	\caption{The table shows the details of different variations (scenes, illumination and weather with different camera angles) present in the FineGPR dataset.}
	\label{table:variabilityfinegpr}
\end{table}
 
 In the equations~\ref{viewpoints},~\ref{backgrounds} and~\ref{weaill}, $p_1$ refers to person 1, $p_2$ refers to person 2, $a_k$, $s_k$ and $w_k$ refers to a fixed viewpoint(angle) k, scene k and weather/illumination k respectively. These observations can be seen in figures~\ref{fig:varvoilin} where the blue plots; corresponding to person 1, are closer to person-1 with a fixed variation 1 as compared to the orange plots that correspond to person 2.
 
 Our study can also help in understanding the scenarios where the selected models may not perform well and the dataset can be updated with these scenarios accordingly.
For instance, adding Gaussian blur to all the images makes the dataset more challenging. Some models (CLIPModel and DinoV2) are more robust to this variation than others (ConvNeXt, ENet and SwinV2) across different scenes, illuminations, weather conditions and viewpoints. Figure~\ref{fig:weatherenetvoilin} shows the results corresponding to the ENet model.
 
 Overall, we find that in non-ambiguous scenarios, the distances can be used to distinguish between different persons in diverse scenes, viewpoints, illumination and weather conditions.
 However, an in depth analysis might be required for similar looking persons.
 
\end{document}